\documentclass[10pt,journal,compsoc]{IEEEtran}

\ifCLASSOPTIONcompsoc
  \usepackage[nocompress]{cite}
\else
  \usepackage{cite}
\fi

\usepackage{epsfig}
\usepackage{graphicx}
\usepackage{amsmath}
\usepackage{amssymb}
\usepackage{graphicx}
\usepackage{amsmath}
\usepackage{amssymb}
\usepackage{algorithm}
\usepackage{algorithmic}
\usepackage{booktabs}
\usepackage{subfigure}
\usepackage{stfloats}
\usepackage{nccmath}
\usepackage{bm}
\usepackage{multirow}
\usepackage{multicol}
\usepackage{array}
\usepackage{url}         
\usepackage{xcolor} 
\usepackage{epsfig} 
\usepackage{colortbl}
\usepackage{arydshln}
\usepackage{array}
\usepackage{booktabs}
\usepackage{threeparttable}
\usepackage[shortlabels]{enumitem}
\usepackage{subfigure}
\usepackage{caption}

\begin{document}

\newcommand{\tabincell}[2]{\begin{tabular}{@{}#1@{}}#2\end{tabular}}
\newcommand{\PreserveBackslash}[1]{\let\temp=\\#1\let\\=\temp}
\newcolumntype{C}[1]{>{\PreserveBackslash\centering}p{#1}}

%
\title{Learning Continuous Face Age Progression: A Pyramid of GANs}
%
%
%
%

\author{Hongyu~Yang,~\IEEEmembership{Student Member,~IEEE,}
        Di~Huang,~\IEEEmembership{Member,~IEEE,}
        Yunhong~Wang,~\IEEEmembership{Senior Member,~IEEE,}
        and~Anil~K.~Jain,~\IEEEmembership{Life~Fellow,~IEEE}
\IEEEcompsocitemizethanks{\IEEEcompsocthanksitem H. Yang, D. Huang, and Y. Wang are with the Beijing Advanced Innovation Center for Big Data and Brain Computing, Beihang University, Beijing 100191, China.

E-mail:\{hongyuyang, dhuang, yhwang\}@buaa.edu.cn
\IEEEcompsocthanksitem Anil K. Jain is with the Department of Computer Science and Engineering, Michigan State University, East Lansing, MI, 48824, USA. 

E-mail: jain@cse.msu.edu}
}

\IEEEtitleabstractindextext{%
\begin{abstract}
The two underlying requirements of face age progression, i.e. aging accuracy and identity permanence, are not well studied in the literature. This paper presents a novel generative adversarial network based approach to address the issues in a coupled manner. It separately models the constraints for the intrinsic subject-specific characteristics and the age-specific facial changes with respect to the elapsed time, ensuring that the generated faces present desired aging effects while simultaneously keeping personalized properties stable. To ensure photo-realistic facial details, high-level age-specific features conveyed by the synthesized face are estimated by a pyramidal adversarial discriminator at multiple scales, which simulates the aging effects with finer details. Further, an adversarial learning scheme is introduced to simultaneously train a single generator and multiple parallel discriminators, resulting in smooth continuous face aging sequences. The proposed method is applicable even in the presence of variations in pose, expression, makeup, etc., achieving remarkably vivid aging effects. Quantitative evaluations by a COTS face recognition system demonstrate that the target age distributions are accurately recovered, and 99.88\% and 99.98\% age progressed faces can be correctly verified at 0.001\% FAR after age transformations of approximately 28 and 23 years elapsed time on the MORPH and CACD databases, respectively. Both visual and quantitative assessments show that the approach advances the state-of-the-art.

\end{abstract}

}

\maketitle

\IEEEdisplaynontitleabstractindextext

%
\IEEEpeerreviewmaketitle

\IEEEraisesectionheading{\section{Introduction}\label{sec:introduction}}
\IEEEPARstart{T}{he} famed portrait 'Afghan girl' gained worldwide recognition when it was featured on the cover of National Geographic Magazine in 1985, whereas the person in the imagery remained anonymous for years until she was identified in 2002 \cite{ALifeRevealed}. In thousands of similar cases of searching for long-lost persons or fugitives, there are usually no more clues than old photos. While human beings, especially forensic artists, can attempt to conceive the aging process on individuals' faces, the output apparently depend on their expertise and state of mind. The computer-aided age progression\footnote{We use aging simulation, aging synthesis, and age progression alternately in this paper.}/regression technique is to aesthetically render a given face image with natural aging/rejuvenating effects. By generating an accurate likeness years prior to or after the reference photo, it facilitates finding lost individuals and suspect identification in law enforcement, and helps guarding vulnerable population against serial offenders. Furthermore, being described as 'half art and half science\cite{H1996applications}', it also benefits anthropometry, biometrics, entertainment, and cosmetology. Studying face age progression is thus of great significance, and this paper focuses on this problem.

\begin{figure}[t] 
\centering
\includegraphics[width=3.5in]{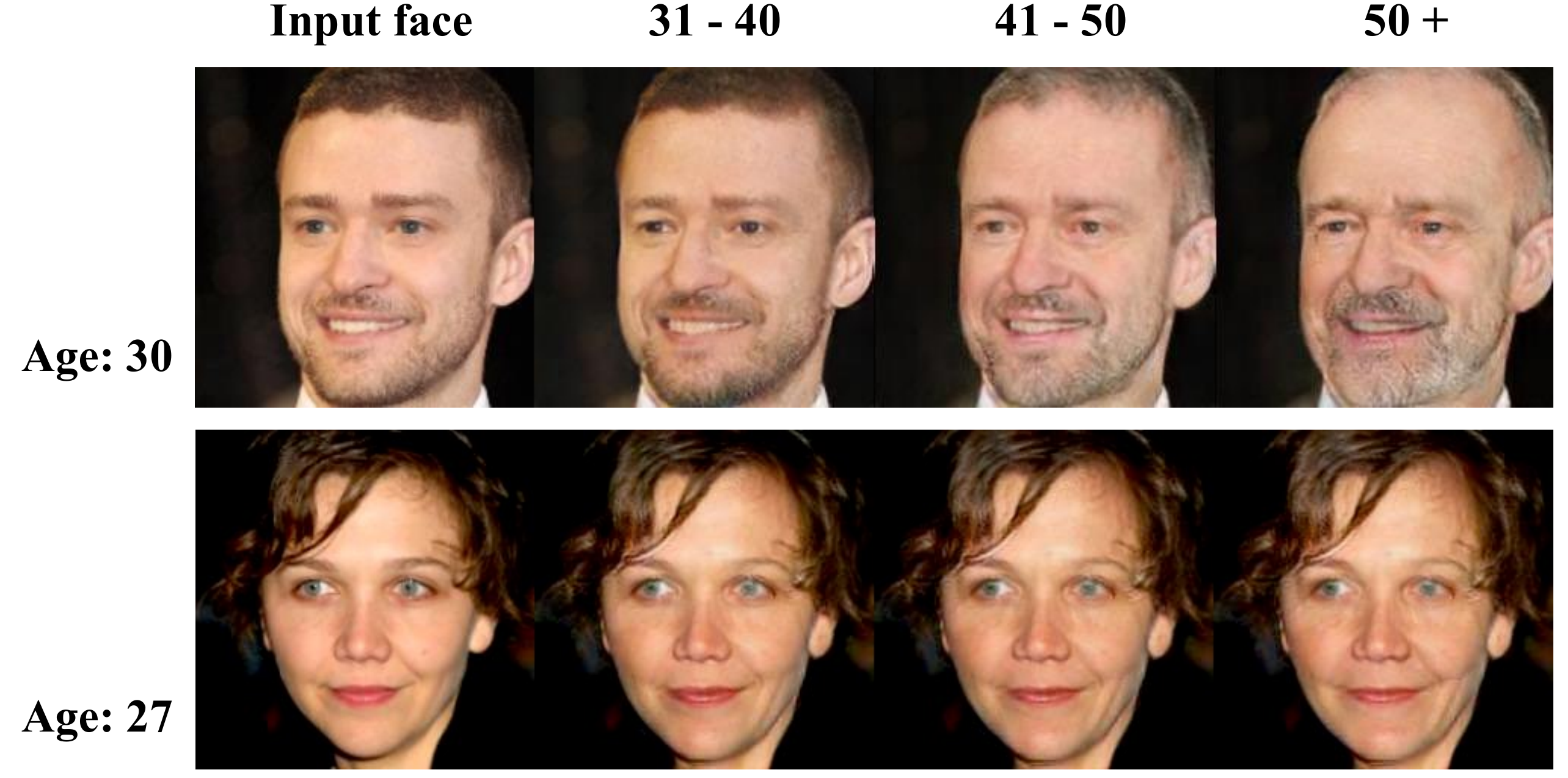}
\caption{Demonstration of our age progression results of two subjects from the CACD aging database (images in the first column are input young faces of two subjects and the others are synthesized older appearances at different age groups).} 
\vspace{-0.2cm}
\end{figure}

Face aging is a process that happens throughout our lives. The intrinsic particularity and complexity of physical aging, the interferences caused by other factors (\emph{e.g.}, PIE variations), and shortage of labeled aging data, collectively make learning face age progression a rather difficult problem. Ever since Pittenger and Shaw\cite{Shaw1975} presented a theory of event perception to simulate the craniofacial growth in 1975, substantial efforts have been made to tackle the challenges of aging simulation, where aging accuracy and identity permanence are commonly acknowledged as the two underlying premises of its success\cite{Suo:Compositional}\cite{Yang:faceAging}\cite{shu2015personalized}\cite{lanitis2008evaluating}. Technological advancements have undergone a gradual transition from computer graphics to computer vision, with deep generative networks now dominating this community.

The pioneering studies on this issue mechanically simulated the profile growth and muscle changes w.r.t. the elapsed time, where crania development theory and skin wrinkle analysis were investigated \cite{todd1980perception}\cite{wu1994plastic}\cite{ramanathan2008modeling}. These methods provided the first insight into face aging synthesis; however, they heavily relied on the empirical knowledge, and generally worked in a complex manner, making them difficult to generalize. Data-driven approaches followed, where age progression was primarily carried out by applying the prototype of aging details to test faces \cite{Suo:Compositional}\cite{Kemelmacher:aging}, or by modeling the dependency between longitudinal facial changes and corresponding ages \cite{park2010age}\cite{Suo:Concatenational}\cite{Wang:tensorAging}. Although obvious signs of aging were well synthesized, their aging functions cannot formulate the complex mechanism accurately enough, limiting the diversity of aging patterns.

Until quite recently, deep generative networks have exhibited a remarkable capability in image generation \cite{gan}\cite{Phillip2016image2image}\cite{taigman2016unsupervised}\cite{DB16c} and have also been utilized for age progression \cite{nhan2016longitudinal}\cite{Wang:Recurrent}\cite{zhang2017age}\cite{nhan2017temporal}. While these approaches render faces with more appealing aging effects and less ghosting artifacts compared to the previous conventional approaches, the problem has not been solved. Specifically, these approaches focus more on modeling face transformation between two age groups, where the age factor plays a dominant role while the identity information plays a subordinate role, with the result that aging accuracy and identity permanence can hardly be simultaneously achieved, in particular for long-term age progression \cite{nhan2016longitudinal}\cite{nhan2017temporal}. Furthermore, they mostly require multiple face images of different ages of the same individual at the training stage, involving another intractable issue, \emph{i.e.} intra-individual aging face sequence collection \cite{Wang:Recurrent}\cite{liu2017aging}. Both the aforementioned facts 
underscore the need of improving the capability of face age progression.

In this study, we propose a novel approach to face age progression, which integrates the advantage of Generative Adversarial Networks (GAN) in synthesizing visually plausible images with prior domain knowledge in human aging. Compared with existing methods in the literature, it is more capable of handling the two critical requirements in age progression, \emph{i.e.} identity permanence and aging accuracy, delivering continuous sequences with more realistic effects. To be specific, the proposed approach uses a Convolutional Neural Network (CNN) based generator to capture target age distributions, and it separately models different face attributes depending upon their changes over time. The training critic thus incorporates the squared Euclidean loss in the image space, the GAN loss that encourages generated faces to be indistinguishable from the age progressed faces in the training set in terms of age, and the identity loss which minimizes the input-output distance by a high-level feature representation embedding personalized characteristics. It ensures that the resulting faces present desired effects of aging while the identity properties remain stable. 

In contrast to the previous techniques that primarily operate on cropped facial areas (usually excluding foreheads) \cite{Yang:faceAging}\cite{Wang:Recurrent}\cite{zhang2017age}\cite{nhan2017temporal}\cite{liu2017aging}, we highlight that synthesis of the entire face is important since the parts of forehead and hair also significantly impact the perceived age. To achieve this and further enhance the aging details, our method leverages the intrinsic hierarchy of deep networks. A discriminator of the pyramid architecture is designed to estimate high-level age-related clues in a fine-grained way. Our approach overcomes the limitations of single age-specific representation and handles age transformation both locally and globally. As a result, more photorealistic imageries are generated (see Fig. 1 for an illustration of aging results). 

Additionally, through an extended GAN structure, consisting of a single generator and multiple parallel discriminators, we can render the input face image to any arbitrary age label and produce a continuous face aging sequence, which tends to support more generalized application scenarios. As the data density of each individual age cluster is jointly considered, the proposed approach does not demand face pairs across two age domains or entire aging sequences of the same person in the training phase as the majority of the counterparts do, thus alleviating the problem of large data collection. 

More concisely, this study makes the following contributions: 

\begin{enumerate}[1)]
\item A novel GAN based method for age progression, which incorporates face verification and age estimation techniques, thereby addressing the issues of aging effect generation and identity cue preservation in a coupled manner;
\item A pyramid-structured discriminator for GAN-based face synthesis, which well simulates both global muscle sagging and local subtle wrinkles;
\item An adversarial learning scheme to simultaneously train a single generator and multiple parallel discriminators, which is able to generate smooth continuous aging sequences even if only faces from discrete age clusters are provided;
\item New validation experiments in addition to existing protocols, including COTS face recognition system based evaluation and robustness assessment to the changes in expression, pose, and makeup. 
\end{enumerate}

A preliminary version of this paper was published in \cite{yang2018AgeProgression}. This paper significantly improves \cite{yang2018AgeProgression} in the following aspects. (i) We extend the model to incorporate the conditional age information; the new model iteratively learns age transformations for diverse target age groups, which simplifies the complex training procedure that requires individual training sessions for different target age groups in \cite{yang2018AgeProgression}. (ii) We extend the model to progressive aging simulation covering any arbitrary age; whereas \cite{yang2018AgeProgression} merely approximates the age distributions of given face sets. (iii) Both face age progression and regression results are refined, along with more extensive evaluations and more comprehensive discussions.

The rest of this paper is organized as follows. Section 2 reviews related work on face age progression. Section 3 details the proposed GAN based aging simulation method. Section 4 displays and analyzes the experimental results on three databases, followed by Section 5 concluding this paper with perspectives.

\begin{table*}[t] \footnotesize
\centering
\caption{Summary of recent representative studies on face age progression}
\begin{threeparttable}
\scalebox{0.92}[0.92]{
\begin{tabular}{cccll}
\toprule
\multirow{2}{*}{Methods} & \multicolumn{2}{c}{~~~~~~~Databases}   & \multirow{2}{*}{~~~~~~~~~~~~~~~Quantitative Evaluations} &  \multirow{2}{*}{~~~~~~~Remarks} \\  
&Training&Testing & & \\
\specialrule{0em}{1pt}{1pt}
\cline{1-5}
\specialrule{0em}{1pt}{1pt}
\multirow{3}{*}{\tabincell{c}{Illumination-aware \\Prototyping \cite{Kemelmacher:aging} }}& \multirow{3}{*}{40K images}& \multirow{3}{*}{FG-NET}&\multirow{3}{*}{\tabincell{l}{Subjective preference votes are higher than that\\of prior work for aging young children on 120\\ aged face pairs. }} &\multirow{2}{*}{\tabincell{l}{Age related high-frequency details \\are smoothed out during computing\\ the templated aging mask.}}\\
&&&&\\
&&&&\\
\cline{1-5}
\specialrule{0em}{1pt}{1pt}

\multirow{5}{*}{\tabincell{c}{Hidden Face Analysis\\ joint \\Sparse Representation\\ \cite{Yang:faceAging} }}  & \multirow{5}{*}{\tabincell{c}{IRIP\\(2,100 images)}}&\multirow{5}{*}{\tabincell{c}{FG-NET, \\MORPH,\\ IRIP}}&   \multirow{5}{*}{\tabincell{l}{Perceived ages of the synthetic faces increase\\along with target ages; Rank-1 recognition rates\\on 20 random selected subjects\tnote{a}~~exceed 70\% for\\the target age cluster of [31-40] on the listed\\databases.}}& \multirow{5}{*}{\tabincell{l}{Aging patterns are linearly modeled.}}\\
&&&&\\
&&&&\\
&&&&\\
&&&&\\
\cline{1-5}
\specialrule{0em}{1pt}{1pt}
\multirow{4}{*}{\tabincell{c}{Recurrent Face Aging\\ \cite{Wang:Recurrent}} }& \multirow{4}{*}{\tabincell{c}{4,371 male\\ images, 6,264 \\female images}}  & \multirow{4}{*}{FG-NET}& \multirow{4}{*}{\tabincell{l}{EER is lower in cross-age face verification on 916\\synthetic pairs\tnote{b}   ~than on original pairs\tnote{c}~; subjective\\preference votes are higher (58.67\%) than that of\\prior work (30.92\%) on 246 aged face pairs.}} & \multirow{4}{*}{\tabincell{l}{Aging sequences are required for\\training the network; testing is \\inflexible.}} \\
&&&&\\
&&&&\\
&&&&\\
\cline{1-5}
\specialrule{0em}{1pt}{1pt}
\multirow{4}{*}{\tabincell{c}{Conditional Adversarial \\Autoencoder (CAAE)\\ \cite{zhang2017age} }}  &\ \multirow{4}{*}{10,670 images}&\multirow{4}{*}{FG-NET} & \multirow{4}{*}{\tabincell{l}{48.38\%~age progressed faces can be verified in human\\based evaluation on 856 synthetic pairs\tnote{b}~; ~subjective\\ preference votes (52.77\%) are higher than that of\\ prior work (28.99\%) on 235 aged face pairs.}}& \multirow{4}{*}{\tabincell{l}{Aging details are inadequate due to \\the insufficient representation ability \\of the adversarial discriminator.}}\\
&&&&\\ 
&&&&\\
&&&&\\
\cline{1-5}
\specialrule{0em}{1pt}{1pt}
\multirow{3}{*}{\tabincell{c}{Temporal Non-Volume \\ Preserving (TNVP) \cite{nhan2017temporal} }}   & \multirow{3}{*}{\tabincell{c}{AGFW(18,685 \\images), 6,437 \\aging sequences}}&\multirow{3}{*}{\tabincell{c}{FG-NET, \\MORPH}}& \multirow{3}{*}{\tabincell{l}{On FG-NET, TAR at 0.01\% FAR is 47.72\% in age\\ invariant face verification including 1M distractors.}}& \multirow{3}{*}{\tabincell{l}{Without identity consistency.}}\\
&&&&\\
&&&&\\
\bottomrule
\specialrule{0em}{1pt}{1pt}  
\end{tabular}}

\begin{tablenotes}
\footnotesize
\item[a] The raw test faces in the Gallery set; their corresponding age-progressed faces in the Probe set.
\item[b] Consisting of an age-progressed face image and a ground-truth image.
\item[c] Consisting of a raw test face image and a ground-truth image.
\end{tablenotes} 
\end{threeparttable}
\vspace{-0.05cm}
\end{table*}

\vspace{-0.1cm}
\section{Related Work}
The published studies on face age progression can be primarily summarized into: (i) empirical knowledge based models, (ii) conventional statistical learning based models, and (iii) deep generative models. In the following, we briefly review these approaches in terms of algorithm, database, and evaluation metrics. 

\subsection{On Algorithm} 
\textbf{I. Empirical knowledge based models.}
Such methods were exploited in the initial explorations of face age progression to simulate the aging mechanisms of cranium and facial muscles. Todd \emph{et al.} \cite{todd1980perception} introduced a revised cardioidal-strain transformation where head growth was modeled in a computable geometric procedure. Based on skin's anatomical structure, Wu \textit{et al.} \cite{wu1994plastic} proposed a 3-layered dynamic skin model to simulate wrinkles. Mechanical aging methods were also incorporated by Ramanathan and Chellappa \cite{ramanathan2008modeling} and Suo  \textit{et al.} \cite{Suo:Concatenational}. Although promising results are reached, it is not so straightforward to generalize these models, as they highly depend on specialized rules and operate in a sophisticated way. 

\textbf{II. Conventional statistical learning based models.} The aging patterns were basically learned from the training faces covering a wide range of ages. Kemelmacher-Shlizerman \emph{et al.} \cite{Kemelmacher:aging} presented a prototype based method, and they further took the illumination factor into account. Wang \emph{et al.} \cite{Wang:tensorAging} built the mapping between corresponding down-sampled and high-resolution faces in a tensor space, and aging details were added on the latter. Yang \emph{et al.} \cite{Yang:faceAging} first settled the multi-attribute decomposition problem, and progression was achieved by transforming only the age component to a target age group. These methods indeed improve the results; however, the aging prototypes or functions cannot accurately fit the aging process, leading to lack of aging diversity. Meanwhile, ghosting artifacts frequently appear in the synthesized faces. 

\textbf{III. Deep generative models.} These methods encode facial variations in terms of age by hierarchically learned deep features for simulation. In \cite{Wang:Recurrent}, Wang \emph{et al.} transformed faces across different ages smoothly by modeling the intermediate transition states in an RNN model. But multiple face images of various ages of each subject were required at the training stage, and the exact age label of the probe face was needed during test, thus greatly limiting its flexibility. Under the framework of conditional adversarial autoencoder (CAAE) \cite{zhang2017age}, facial muscle sagging caused by aging was simulated, whereas only rough wrinkles were rendered mainly due to the insufficient representation ability of the discriminator. With the Temporal Non-Volume Preserving (TNVP) aging approach \cite{nhan2017temporal}, the short-term age progression was accomplished by mapping the data densities of two consecutive age groups with ResNet blocks \cite{he2016deepresidual}, and the long-term aging synthesis was finally reached by a chaining of short-term stages. Its major weakness, however, was that it merely considered the probability distribution of a set of faces without any individuality information. As a result, the synthesized faces in a complete aging sequence varied a lot in color, expression, and even identity.

\subsection{On Data}
Two databases have been widely used in face age progression, namely FG-NET \cite{fgnet} and MORPH \cite{morph}, and they have greatly facilitated progress in the community. Since FG-NET only contains 1,002 face images from 82 subjects, it is commonly used in the test phase only \cite{Kemelmacher:aging}\cite{shu2015personalized}\cite{zhang2017age}. The MORPH mugshot database is relatively large, which consists of more than 50K images from over 13K subjects; however, the intensive image acquisition time and limited number of images per subject (the average time lapse per subject is only 1.62 years and the average number of images per subject is 4.03) does not make MORPH suitable for approaches requiring the long-term aging sequence from an individual. As a result, some studies, \emph{e.g.}, \cite{shu2015personalized}\cite{nhan2017temporal}\cite{liu2017aging}, make use of their private databases or combine the existing ones for training. Besides these databases, Cross-Age Celebrity Dataset (CACD)\cite{2015cacd} also incorporates longitudinal aging variations, containing 163,446 images 'in the wild' from 2,000 celebrities, which can be further exploited for modeling face aging.

\subsection{On Evaluation Metrics}

\begin{figure*}[t]
\centering 
\includegraphics[width=0.98\textwidth]{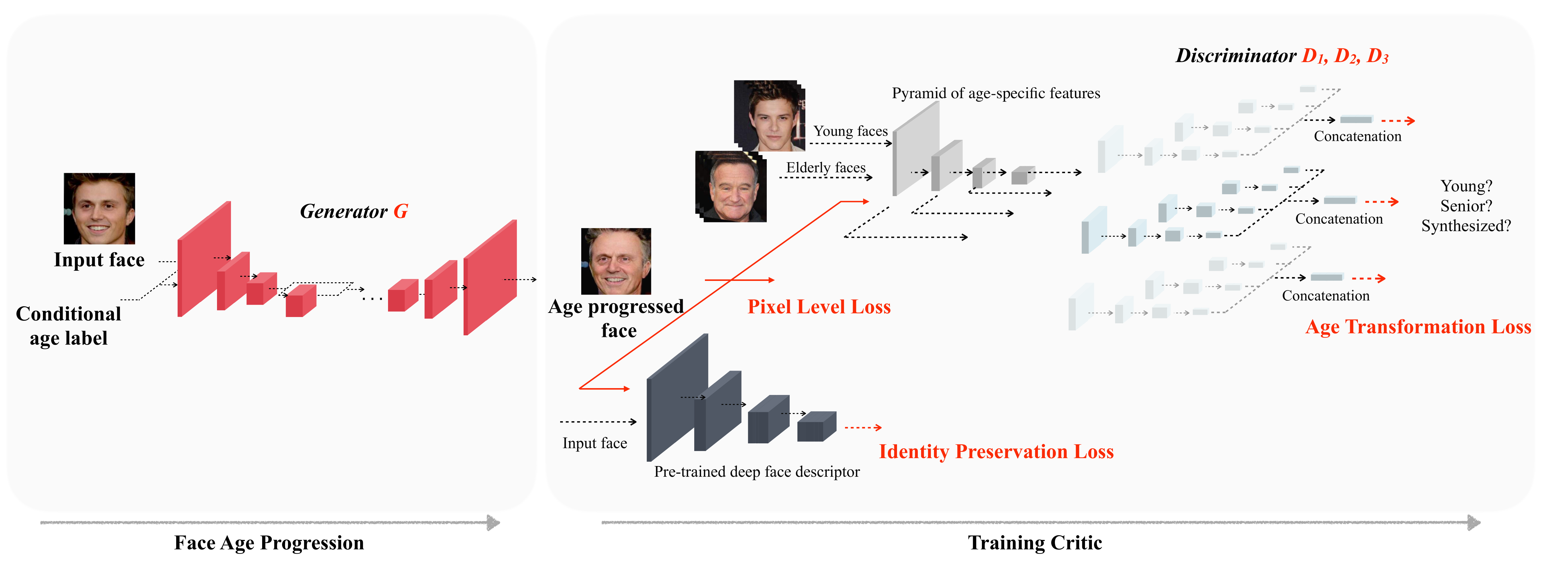}
\caption{Framework of the proposed age progression method. A CNN based generator $G$ learns the age transformation. It takes a younger face and the conditional target age label as inputs, and outputs a corresponding aged likeness. The training critic incorporates the squared Euclidean loss in the image space, the GAN loss that encourages generated faces to be indistinguishable from the training elderly faces in terms of age, and the identity preservation loss minimizing the input-output distance in a high-level feature representation which embeds the personalized characteristics.}
\vspace{-0.2cm}
\end{figure*}

Early research validated the proposed methods by comparative visualizations between a few example face images and their corresponding age progressed results. While visual inspection is useful, it is subjective and not specific. The two criteria for aging model evaluation, \emph{i.e.} accuracy of aging and preservation of identity, were then proposed by Lanitis \cite{2008LanitisComparative} and refined by Suo \emph{et al.} \cite{Suo:Compositional}. They were not only intuitively reasonable, but also quantitatively feasible. Therefore, in the subsequent studies \cite{Yang:faceAging}\cite{Wang:Recurrent}\cite{zhang2017age}\cite{nhan2017temporal}, face recognition/verification on the generated faces was often conducted for performance assessment. To the best of our knowledge, however, such evaluations were mostly performed on a small number of faces, and the data (images or subjects) used were often arbitrary and varied from one study to another. Furthermore, evaluations on the perceived/estimated age\footnote{Perceived age: the individual age gauged by human subjects from the visual appearance. Estimated age: The individual age recognized by machine from the visual appearance\cite{Fu:survey}.} of the simulated faces have received limited attention.

Table 1 presents a summary of the recent representative studies. Despite this progress, these approaches either cannot fully address simulation accuracy or identity permanence, or neither; and the evaluation metrics have some limitations. Apart from these issues, generating rich texture still remains a challenge in many scenarios of face synthesis, especially for the specific task of age progression, where visually convincing wrinkles and age-related details are extremely essential to accurate perceived age and authenticity.

Our study makes use of the ability of image generation by GAN and presents a different but effective method, where the age-related GAN loss is adopted for aging modeling, the individual-dependent critic is used to keep the identity cue stable, and a multi-pathway discriminator architecture is further applied to refine aging detail generation. This solution is more powerful in dealing with the core issues of age progression, \emph{i.e.} age accuracy and identity preservation. Meanwhile, it is able to produce continuous face aging sequences without any strong assumptions on training data. 
Additionally, it shows the robustness to expression, pose, and makeup variations.

\section{Method}

\subsection{Overview}
A classic GAN contains a generator $G$ and a discriminator $D$, which are iteratively trained via an adversarial process. The generative function $G$ tries to capture the underlying data density and confuse the discriminative function $D$, while the optimization procedure of $D$ aims to achieve the distinguishability and distinguish the natural face images from the fake ones generated by $G$. Both $G$ and $D$ can be approximated by neural networks, \textit{e.g.}, Multi-Layer Perceptron (MLP). The risk function of optimizing this minimax two-player game can be written as:

\begin{equation}
\begin{aligned}
\mathcal{V}(D, G) = &\min \limits_{G} \max \limits_{D}  \mathbb{E}_{x\sim P_{data}} {\rm log}[D(x)]\\
 &+ \mathbb{E}_{z\sim P_{z}} {\rm log} [1-D(G(z))] 
\end{aligned}
\end{equation}
\noindent where $z$ is a noise sample from a prior probability distribution $P_{z}$, and $x$ denotes a real face image following a certain distribution $P_{data}$. On convergence, the distribution of the synthesized images $P_{G}$ is equivalent to $P_{data}$. 

Recently, the conditional GANs (cGANs) have been actively studied, where the generative model $G$ approximates the dependency of the pre-images (or controlled attributes) and their corresponding targets. cGANs have shown promising results in video prediction \cite{mathieu2015deep}, text to image synthesis \cite{reed2016generative}, image-to-image translation \cite{Phillip2016image2image}\cite{zhu2017unpaired}, \emph{etc}. In our case, a CNN based generator takes the younger face image and the target age label (or a target age group) as inputs, and synthesizes an elder face image conditioned on them.

In order to accurately achieve the aging effects while simultaneously maintaining the person-specific information, a compound training critic is exploited in the offline phase, which incorporates the traditional squared Euclidean loss in the image space, the GAN loss that encourages generated faces to be indistinguishable from the training elderly faces in terms of age, and the identity loss minimizing the input-output distance in a high-level feature representation embedding the personalized characteristics. Note, in adversarial training, a pyramid-structured discriminator is specially designed to refine facial detail generation. The classic GAN model is extended, where one single generator is utilized along with a specific number of parallel discriminators, in order to flexibly steer the age transformation to diverse target age labels. The adversarial training scheme in our method does not only contribute to the real-fake level classification, but also guides the model to converge towards target age distributions. See Fig. 2 for an overview, and we detail the method in the subsequent sections.

\subsection{Generator}

The proposed model enables one-step aging simulation. Synthesizing age progressed faces only requires a forward pass through the generator $G$. The generative network is a cascade of encoder and decoder. With the input young face $x$ and the target age label (or age range) $c$, it first exploits multiple stacked convolutional layers to encode them to a latent space, capturing the facial properties that tend to be stable w.r.t. the elapsed time, followed by four residual blocks \cite{he2016deepresidual} modeling the common structure shared by the input and output faces, similar to the settings in \cite{johnson2016perceptual}. Age transformation to a target image space is finally achieved by three fractionally-strided convolutional layers, yielding the age progression result $y$ conditioned on the inputs, $G(x,c) \rightarrow y$. Rather than using the max-pooling and upsampling layers to calculate the feature maps, the $3 \times 3$ convolution kernels with a stride of 2 are employed here, ensuring that every pixel contributes and the adjacent pixels transform in a synergistic manner. All the convolutional layers are followed by Instance Normalization and ReLU non-linearity activation. Paddings are added to the layers to make the input and output have exactly the same size. A total variation regularization layer is stacked to the end of $G$ to reduce the spike artifacts. The detailed generator architecture is shown in \textit{Section 4.2}.

\subsection{Adversarial Learning}
To ensure the generated face images present proper aging effects, we adopt the adversarial learning mechanism. We first introduce how it is exploited to achieve age transformation to a domain corresponding to a specific target age cluster and then illustrate how this method is generalized to simulate the progressive aging procedure.
\subsubsection{Aging Modeling}
The system critic incorporates the prior knowledge of the data density of the faces from the target age cluster $c$, and a discriminative network $D$ is thus introduced, which outputs a scalar representing the probability that $x$ comes from the data, $x \rightarrow D(x)$. We denote the distribution of young faces as $x\sim P_{young}$ and the distribution of the generated faces as $G(x, c)\sim P_{G}$.  $P_{G}$ is supposed to be equivalent to the distribution $P_{old}$ of the target age cluster when age transformation is learned. Assuming that we follow the classic GANs \cite{gan}, where a binary cross entropy classifier is used, then the learning process amounts to minimizing the following loss defined over $G$ and $D$:

\begin{equation}
\begin{aligned}
\mathcal{L}_{GAN\_D} = & - \mathbb{E}_{x \sim P_{young}, c} {\rm log}[1 - D(G(x, c))]\\
& - \mathbb{E}_{x \sim P_{old}} {\rm log}[D(x)] 
\end{aligned}
\end{equation}

It is always desirable that $G$ and $D$ converge coherently; however, $D$ frequently achieves the distinguishability faster in practice and feeds back vanishing gradients for $G$ to learn, since the JS divergence is locally saturated. As analyzed in some recent studies, \emph{i.e.} the Wasserstein GAN \cite{wgan}, the Least Squares GAN \cite{mao2016least}, and the Loss-Sensitive GAN \cite{lsgan}, the most fundamental issue lies in how exactly the distance between sequences of probability distributions is defined. Here, we use the least squares loss substituting for the negative log likelihood objective, which penalizes the samples depending on how close they are to the decision boundary in a metric space, minimizing the Pearson $\mathcal{X}^{2}$ divergence. To achieve more evident and vivid age-specific facial details, both the actual young faces and the generated age-progressed faces are fed into $D$ as negative samples while the true elderly face of age range $c$ as positive ones. Accordingly, the training process alternately minimizes the following:

\begin{figure}[t] 
\centering
\includegraphics[width=3.2in]{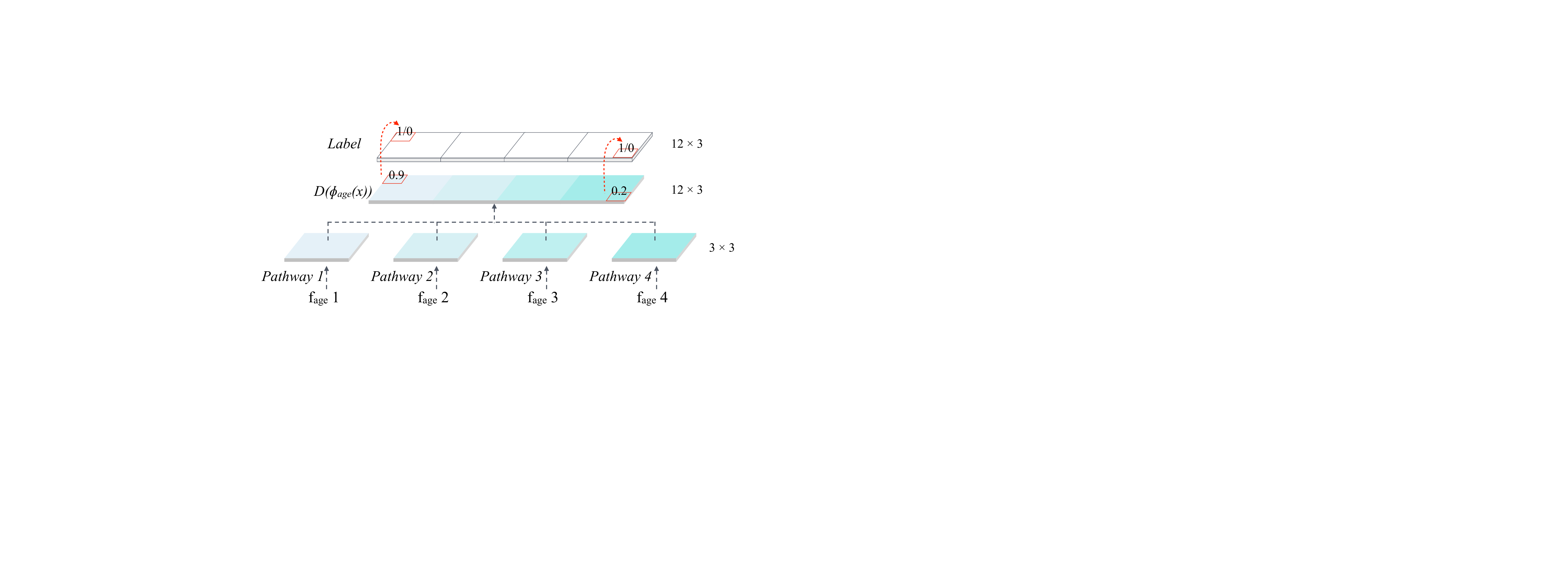}
\caption{The scores of four pathways are finally concatenated and jointly estimated by the discriminator $D_{i}$ ($D_{i}$ is an estimator rather than a classifier; the $Label$ does not need to be a single scalar).} 
\end{figure}

\begin{figure*}[t]
\centering 
\includegraphics[width=0.9\textwidth]{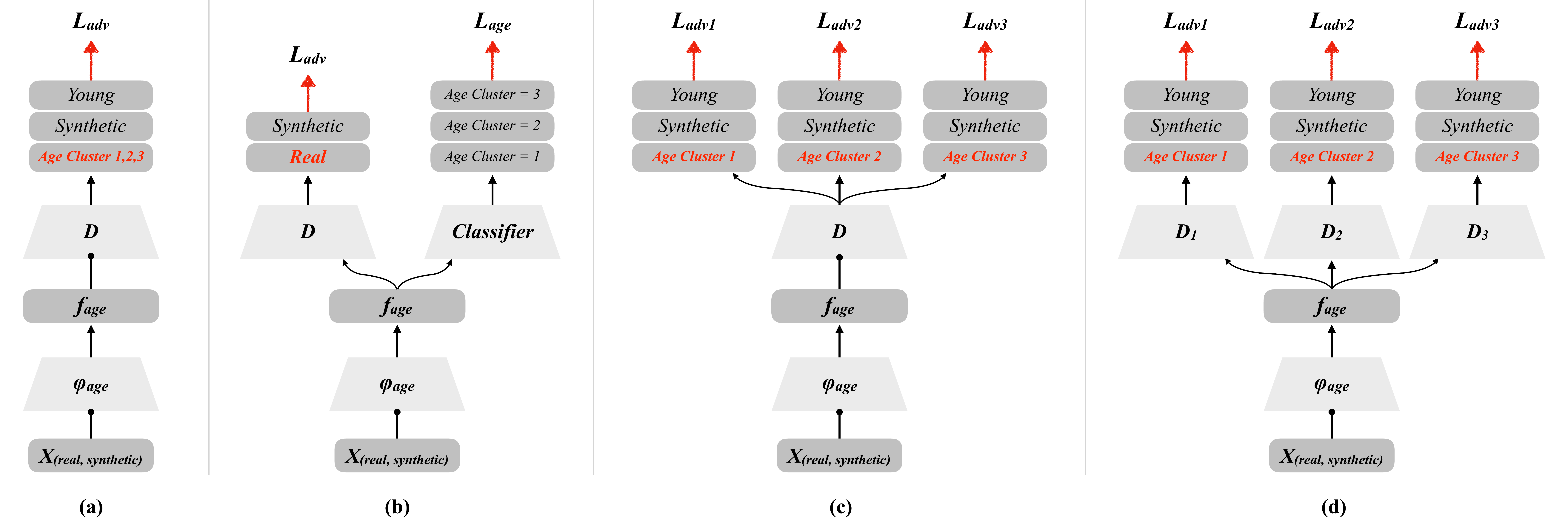}
\caption{Comparison of several extensions of the original adversarial training scheme presented in Section 3.3.1. (a) The general pipeline of original adversarial training; (b) Auxiliary Classifier GAN for progressive aging modeling; (c) $D$ is extended to output $k$ logits handling the degree of aging; and (d) A specific number of discriminators are simultaneously trained to steer the age transformation to different age domains.}
\vspace{-0.25cm}
\end{figure*}

\begin{equation}
\mathcal{L}_{GAN\_G} = \mathbb{E}_{x\sim P_{young}, c}H(1,D(\phi_{age}(G(x, c)))) 
\end{equation} 

\begin{equation}
\begin{aligned}
\mathcal{L}_{GAN\_D} = &~\mathbb{E}_{x\sim P_{young}, y\sim P_{old},c } H([1, 0, 0],
\\&[D(\phi_{age}(y)),D(\phi_{age}(G(x, c))),D(\phi_{age}(x))])
\end{aligned}
\end{equation}
\noindent where $H$ indicates the least squares distance. Note, in (3) and (4), a function $\phi_{age}$ bridges $G$ and $D$, which is especially introduced to extract age-related features conveyed by faces, as shown in Fig. 2. Considering that human faces of diverse age groups share a common configuration and similar texture properties, a feature extractor $\phi_{age}$ is thus exploited independently of $D$, which outputs high-level feature representations to make the generated faces more distinguishable from the true elderly faces in terms of age. In particular, $\phi_{age}$ is pre-trained for a multi-label classification task of age estimation with the VGG-16 structure \cite{simonyan2014very}, and after convergence, we remove the fully connected layers and integrate it into the framework. Further, since natural images exhibit multi-scale characteristics and along the hierarchical architecture, $\phi_{age}$ captures the properties gradually from exact pixel values to high-level age-specific semantic information, this study leverages the intrinsic pyramid hierarchy. The pyramid facial feature representations are jointly estimated by $D$ at multiple scales, handling aging effect generation in a fine-grained way.

\begin{figure}[t]
\centering 
\includegraphics[width=3.5in]{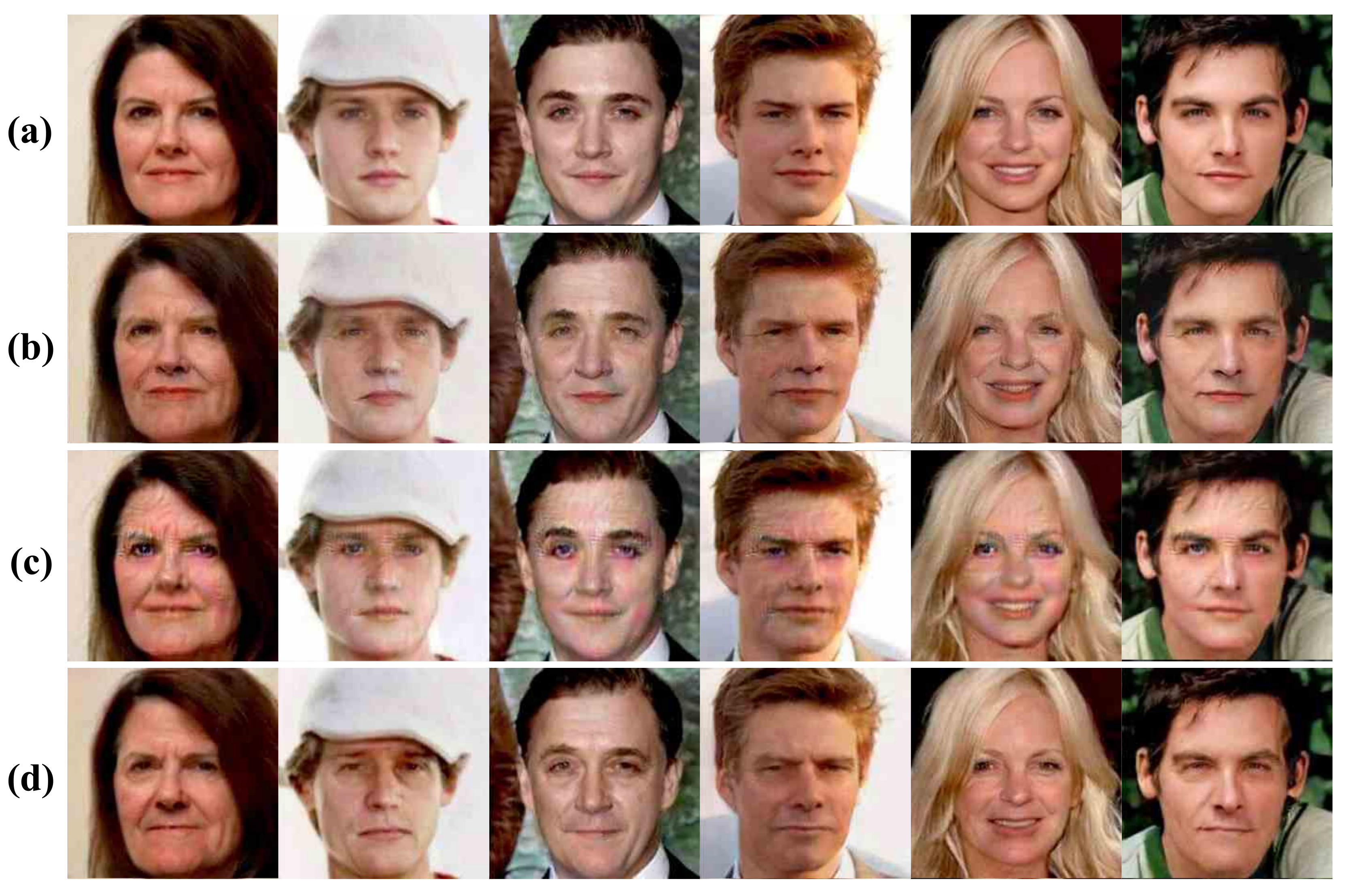}
\caption{Comparison of the aging results achieved by different extensions of the original adversarial training scheme. The fist image in each column is the input face image and the subsequent 3 images are the age progressed visualizations achieved by the methods demonstrated in Figs. 4(b), 4(c), and 4(d), respectively.}
\vspace{-0.25cm}
\end{figure}

The outputs of the 2nd, 4th, 7th and 10th convolutional layers of $\phi_{age}$ are used. They pass through the pathways of $D$ and finally result in a concatenated 12 $\times$ 3 representation. The 'label' applied here is thus a tensor of the same size rather than a single scalar, fulfilled with ones (for the positive face samples) or zeros (for the negative face samples). The least squares loss is minimized using the full feature representation to jointly estimate the 4 pathways, as illustrated in Fig. 3. In $D$, all convolutional layers are followed by Batch Normalization and LeakyReLU activation except the last one. The detailed discriminator architecture is shown in \textit{Section 4.2}.

\subsubsection{Progressive Aging Modeling}
As face aging is a dynamic long-term procedure, we attempt to synthesize progressive changes w.r.t the elapsed time and generate complete aging sequences. Under the GAN framework, a common practice is to leverage the age labels, add an auxiliary classifier (on top of $D$ or parallel to $D$), and impose the age classification loss when optimizing both $G$ and $D$. The adversarial part of the original aging model shown in Fig. 4 (a) is thus extended to that in Fig. 4 (b). Such a variant has indeed been shown to be effective in handling the data with high variability and improving the quality of generated samples \cite{Salimans2016}\cite{2017acgan}. Forcing the proposed method to perform additional age classification, however, does not boost the performance of the core task in this study, \emph{i.e.} aging effect synthesis, and this claim is supported both mathematically and experimentally. To be specific, an AC-GAN can be viewed as a GAN model with a hierarchical classifier \cite{zhou2018amgan}, and the objective is formulated as:

\begin{small}
\begin{equation}
\begin{aligned}
\mathcal{L}_{GAN^{ac}\_G}&=\mathbb{E}_{x\sim P_{young}, c} H(1,D(\phi_{age}(G(x, c))))
\\& + \mathbb{E}_{x\sim P_{young}, c} H'(v(c),C(G(x,c)))\\
\end{aligned}
\end{equation} 

\end{small}
\begin{small}
\begin{equation}
\begin{aligned}
\mathcal{L}_{GAN^{ac}\_D}&=\mathbb{E}_{x\sim P_{data}, c} H([1,0],[D(\phi_{age}(x)), D(\phi_{age}(G(x, c)))])
\\& + \mathbb{E}_{x\sim P_{data}, c} H'(v(c),C(x))
\end{aligned}
\end{equation} 
\end{small}
\noindent where $C$ denotes the auxiliary classifier, $H'$ indicates the cross-entropy, and $v(c) = [v_{1} (c), ..., v_{k} (c)]$ with $v_{i} (c) = 1$ if $i = c$; otherwise $v_{i} (c) = 0$. The hierarchical connection of adversarial training and classification inevitably brings the issue that the former is actually missing when optimizing the latter. As for our task of face age progression, $D$ only works at the real-fake level, but does not control the aging degree. As shown in Fig. 5, with the faces in the first row as inputs, we obtain the corresponding age-progressed faces in the second row. While the age-related facial changes, \textit{e.g.} wrinkles around the eyes, indeed emerge, they are not as natural as the real faces, which confirms the above analysis.

As the adversarial constraint is the key to guarantee the convergence $P_{G} = P_{old}$, the non-hierarchical structure as shown in Fig. 4 (c) can be further considered, where the discriminator $D$ outputs $k$ logits indicating the age range that $x$ belongs to. Theoretically, such adversarial training covers all age clusters; however, using a single discriminator is inadequate to accurately model the complex distribution of multiple age domains, and the networks are more likely to suffer mode collapse. See the examples of aging photos shown in Fig. 5(c). The resulting images can still be recognized as faces, but they are mainly credited to the pixel-level constraint and identity preservation loss that will be illustrated in the next subsection. 

To sufficiently leverage the intrinsic domain transfer ability of GANs, different from original adversarial learning and its two major variants (as in Figs. 4 (a), (b), and (c), respectively), we replace the discriminator in (4) with $k$ class-wise discriminators, as shown in Fig. 4 (d). Assigning each specific target age cluster a unique discriminator $D_{i}$, the objectives of the extended model can be finally formulated as: 

\begin{equation}
\mathcal{L}_{GAN\_G_{i}} = \mathbb{E}_{x\sim P_{young}, c}H(1,D_{i}(\phi_{age}(G(x, c_{i})))) 
\end{equation}

\begin{equation}
\begin{aligned}
\mathcal{L}_{GAN\_D_{i}} = &~\mathbb{E}_{x\sim P_{young}, y\sim P_{old},c } H([1, 0, 0],
\\&[D_{i}(\phi_{age}(y)),D_{i}(\phi_{age}(G(x, c_{i}))),D_{i}(\phi_{age}(x))])
\end{aligned}
\end{equation}

The discriminators synergistically guide the generator $G$ to learn age transformations to the domains associated with label $c$. The simulation results are displayed in Fig. 5 (d), in which the aging effects are more evident and closer to the natural images. 

Although the faces are manually divided into a number of discrete clusters along the timeline, the data are intrinsically connected since face aging is an accumulation of changes over time. Based on the latent manifold assumption of images\cite{Tenenbaum2000}, we further make an inference that aging procedure is a smooth transformation lying in a manifold, and adding sufficient local constraints on the key points in the temporal aspect probably enables the method to achieve globally longitudinal aging. In this case, the method not only accurately models the age distributions that independently presented to the discriminators at the training period, but also successfully simulates continuous aging sequences covering any age label at the test phase.

\subsection{Identity Preservation}
One core issue of face age progression is keeping person-dependent properties stable. Therefore, we incorporate the associated constraint by measuring the input-output distance in a proper feature space, which is sensitive to the identity change while relatively robust to other variations. Specifically, the network of \textit{deep face descriptor} \cite{Deepface} is utilized, denoted as $\phi_{id}$, to encode the personalized information and further define the identity loss function. $\phi_{id}$ is trained with a large face dataset containing millions of face images from thousands of individuals\footnote{The face images are collected via the Google Image Search using the names of 5K celebrities, purified by automatic and manual filtering.}. It is originally bootstrapped by recognizing $N = 2,622$ unique individuals; and the last classification layer is then removed and $\phi_{id}(x)$ is tuned to improve the capability of verification in the Euclidean space using a triplet-loss training scheme. In our case, $\phi_{id}$ is clipped to 10 convolutional layers, and the identity loss is formulated as:

\begin{equation}
\mathcal{L}_{identity} = \mathbb{E}_{x\sim P_{young}, c} ~d(\phi_{id}(x),\phi_{id}(G(x, c)))
\end{equation}
\noindent where $d$ is the squared Euclidean distance between feature representations. For more implementation details of \textit{deep face descriptor}, please refer to \cite{Deepface}.

\subsection{Objective}

Besides the specially designed age-related GAN critic and the identity permanence penalty, a pixel-wise L2 loss in the image space is also adopted for further bridging the input-output gap, \textit{e.g.,} the color aberration, which is formulated as:

\begin{equation}
\mathcal{L}_{pixel} =  \mathbb{E}_{x\sim P_{young}, c} ~\frac{1}{W \times H \times C}\|G(x, c) - x\|_{2}^{2}
\end{equation}
\noindent where $W$, $H$, and $C$ correspond to the image shape. Meanwhile, we make use of the total variation regularizer loss encouraging the spatial smoothness, by stacking a TV regularization layer to the end of $G$ as in \cite{johnson2016perceptual}. Finally, the system training loss can be written as:

\begin{equation}
\mathcal{L}_{G} = \lambda_{a} \sum_{i=1}\mathcal{L}_{GAN\_G_{i}} + \lambda_{p}\mathcal{L}_{pixel} +  \lambda_{i}\mathcal{L}_{identity} +  \lambda_{t}\mathcal{L}_{tv}
\end{equation}

\begin{equation}
\mathcal{L}_{D_{i}} = \mathcal{L}_{GAN\_D_{i}}
\end{equation}
\indent We train the generator $G$ and the discriminators $D_{i}~( i = 1, 2, ..., k)$ alternately until optimality, and finally $G$ learns the desired age transformation and $D_{i}$ becomes a reliable estimator.

\begin{table*}[t] 
\centering
\caption{Statistics of face aging databases used for evaluation}
\setlength{\tabcolsep}{2.5mm}
\begin{tabular}{ccccccc}
\toprule
\textbf{Database}&  \textbf{ \tabincell{c}{Number of\\images}}   & \textbf{ \tabincell{c}{Number of\\subjects}}    & \textbf{\tabincell{c}{ Number of \\images per subject }}  &   \textbf{\tabincell{c}{Time lapse\\per subject (years) }} & \textbf{\tabincell{c}{Age span \\ (years old)}} & \textbf{\tabincell{c}{Average age \\(years old) }} \\  
\cline{2-7}
\specialrule{0em}{1pt}{1pt}
MORPH \cite{morph} & 52,099 & 12,938  & 1 - 53 (avg. 4.03) & 0 - 33 (avg. 1.62) & 16 - 77  & 33.07\\
CACD \cite{2015cacd} & 163,446 & 2,000  & 22 - 139 (avg. 81.72) & 7 - 9 (avg. 8.99) & 14 - 62 & 38.03\\
FG-NET \cite{fgnet} & 1,002 & 82 & 6 - 18 (avg. 12.22) & 11 - 54 (avg. 27.80) & 0 - 69 & 15.84 \\
\bottomrule
\end{tabular}
\end{table*}

\begin{figure*}[t]
\centering 
\includegraphics[width=0.92\textwidth]{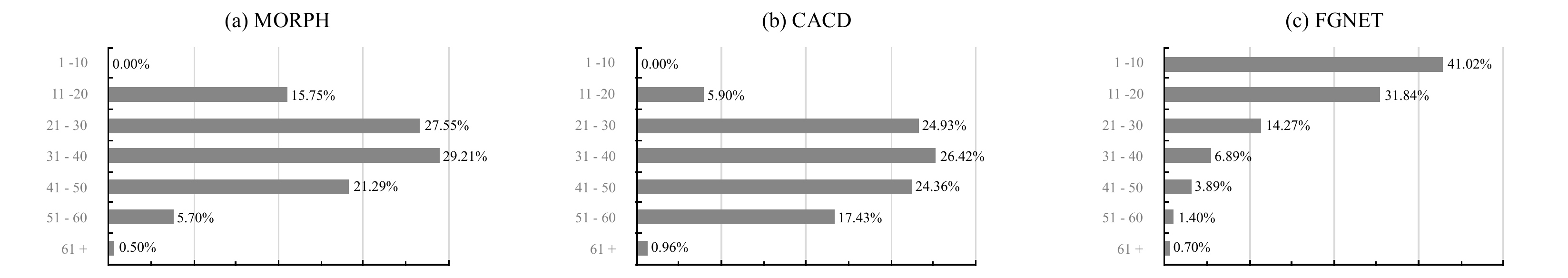}
\caption{Age distributions of (a) MORPH, (b) CACD, and (c) FGNET.}
\end{figure*}

\vspace{-0.05cm}
\section{Experimental Results}
To validate the proposed age progression approach, we carry out extensive experiments and make fair comparison to the state of the art counterparts. The face databases, implementation details, and synthesized results are presented in the subsequent.

\vspace{-0.1cm}
\subsection{Databases}
There are three databases used in the experiments, namely MORPH \cite{morph}, Cross-Age Celebrity Dataset (CACD) \cite{2015cacd}, and FG-NET \cite{fgnet} .

An extension of the \textbf{MORPH} aging database contains 52,099 color images with near-frontal pose, neutral expression, and uniform illumination (minor pose and expression variations sometimes occur). The subject age ranges from 16 to 77 years old, with the average age being approximately 33. The longitudinal age span of a subject varies from 46 days to 33 years. \textbf{CACD} is a public dataset \cite{2015cacd} collected via the Google Image Search, containing 163,446 face images of 2,000 celebrities across 10 years, with age ranging from 14 to 62 years old. The dataset has the largest number of images with age changes, showing variations in pose, illumination, expression, \emph{etc.}, with less controlled acquisition than MORPH. We mainly use MORPH and CACD  for training and validation. \textbf{FG-NET} is also a very popular database for evaluation of face aging methods. As its images are inadequate to train the proposed deep model, we only adopt it for testing to make comparison with prior work. More properties of these databases are given in Table 2 and Figure 6.

\vspace{-0.1cm}
\subsection{Implementation Details}
Allowing for the fact that the number of faces older than 60 years old is quite limited in both training databases of MORPH and CACD, and neither contains images of children, we only perform adult aging. We follow the time span of 10 years for each age cluster as reported in many previous studies \cite{Suo:Compositional}\cite{Yang:faceAging}\cite{Wang:Recurrent}\cite{zhang2017age}\cite{nhan2017temporal}, and apply age progression on the faces below 30 years old, synthesizing a sequence of age-progressed renderings when they are in their 30s, 40s, and 50s.  Prior to feeding the images into the networks, the faces are aligned using the eye locations provided by the datasets themselves (FG-NET, CACD) or detected by the online face analysis API of Face++ \cite{facepp} (MORPH). Excluding those images undetected in MORPH and that of children in FG-NET, 489,~163,446, and 51,699 imageries from the three datasets are finally adopted, respectively. A face image is cropped to 224 $\times$ 224 pixels; concatenating the conditional age, they form a tensor representation of $224 \times 224 \times (3+k)$ as the generator input, where $k$ denotes the number of target age clusters and it is set to 3 in our experiments.

The architectures of the networks $G$ and $D_{i}~(i= 1, 2, 3)$ are shown in Tables 3 and 4. For both training datasets, the trade-off parameters $\lambda_{p}$, $\lambda_{a}$, $\lambda_{i}$, and $\lambda_{t}$ are empirically set to 0.10, 1000.00, 0.005, and $1e^{-6}$, respectively. At the training stage, we use Adam with the learning rate of $1\times {10}^{-4}$ and the weight decay factor of $0.5$ for every $2,000$ iterations. We (i) update the three discriminators alternatively at every iteration, (ii) use the age-related and identity-related critics at every generator iteration, and (iii) employ the pixel-level critic for every 15 generator iterations. The networks are trained with a batch size of 4 for $150,000$ iterations in total, which takes around 25 hours on a GTX 1080Ti GPU.

We comprehensively evaluate the proposed age progression method in the following layers: (I) face aging simulation; (II-A) visual fidelity analysis; objective evaluations on (II-B) accuracy of aging and (II-C) preservation of identity; (II-D) ablation study; and (II-E) comparison with state-of-the-art.

\begin{table}[t] 
\centering
\caption{Generator architecture}
\label{generator}
\begin{tabular}{ccccc}
\toprule
Layer&Kernel &Stride  &Padding &Activation Size\\ 
\midrule
Conv.& 9 $\times$ 9 & 1 & 4 & 32 $\times$ w $\times$ h  \\
Conv.$\downarrow$ & 3 $\times$ 3 & 2 & 1 & 64 $\times$ w/2 $\times$ h/2 \\
Conv.$\downarrow$ & 3 $\times$ 3 & 2 & 1 & 128 $\times$ w/4 $\times$ h/4\\
 Res.& 3 $\times$ 3& 1 & 2 & 128 $\times$ w/4 $\times$ h/4\\
 Res.& 3 $\times$ 3 & 1 & 2 & 128 $\times$ w/4 $\times$ h/4\\
 Res.& 3 $\times$ 3 & 1 & 2 & 128 $\times$ w/4 $\times$ h/4\\
 Res.& 3 $\times$ 3 & 1 & 2 & 128 $\times$ w/4 $\times$ h/4\\
 De-conv.$\uparrow$& 3 $\times$ 3 & 2 & 1 & 64 $\times$ w/2 $\times$ w/2\\
 De-conv.$\uparrow$& 3 $\times$ 3 & 2 & 1 & 32 $\times$ w $\times$ h\\
 De-conv.& 9 $\times$ 9 & 1 & 4 & 3 $\times$ w $\times$ h\\
\bottomrule
\end{tabular}
\end{table}

\begin{table}[t]
\centering
\caption{Discriminator architecture}
\label{discriminator}
\begin{threeparttable}
\setlength{\tabcolsep}{2.6mm}
\begin{tabular}{ccccc}
\toprule
Pathway & \multicolumn{1}{c}{1} & \multicolumn{1}{c}{2}&\multicolumn{1}{c}{3} &\multicolumn{1}{c}{4} \\

\midrule
Input & \multicolumn{1}{c}{512} & \multicolumn{1}{c}{256} & \multicolumn{1}{c}{128} &\multicolumn{1}{c}{64} \\

\midrule
\multirow{6}{*}{Layers\tnote{*}} &   &   &  & conv-128 \\
&  &  & conv-256 & conv-256 \\
&  & conv-512  & conv-512 & conv-512 \\
& conv-512 & conv-512  & conv-512 & conv-512 \\
& conv-512 & conv-512  & conv-512 & conv-512 \\
& conv~~-~~1 & conv~~-~~1  & conv~~-~~1 & conv~~-~~1 \\
\bottomrule
\end{tabular}
\begin{tablenotes}
\footnotesize
\item[*] Layers are denoted as: conv - $<$output$>$; \\kernel = 4, stride = 2, padding = 1
\end{tablenotes} 
\end{threeparttable}
\vspace{-0.2cm}
\end{table}

\begin{figure*}[t] 
\centering
\includegraphics[width=1.0\textwidth]{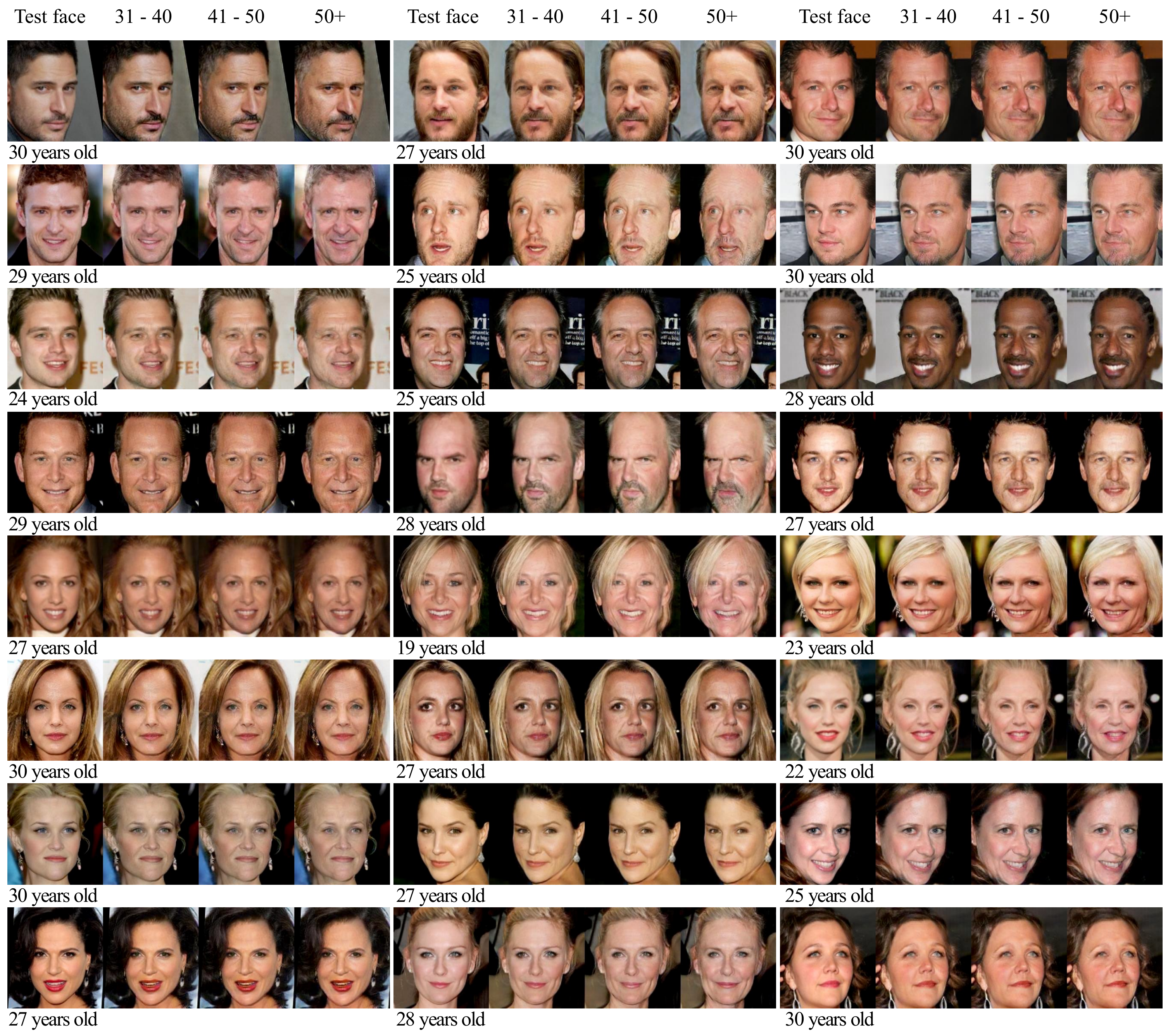}
\vspace{-0.3cm}
\caption{Aging effects obtained on the CACD databases for 24 different subjects. The first image in each panel is the original face image and the subsequent 3 images are the age progressed visualizations for that subject in the [31- 40], [41-50] and 50+ age clusters.} 
\vspace{-0.2cm}
\end{figure*}

\begin{figure*}[t] 
\centering
\includegraphics[width=1.0\textwidth]{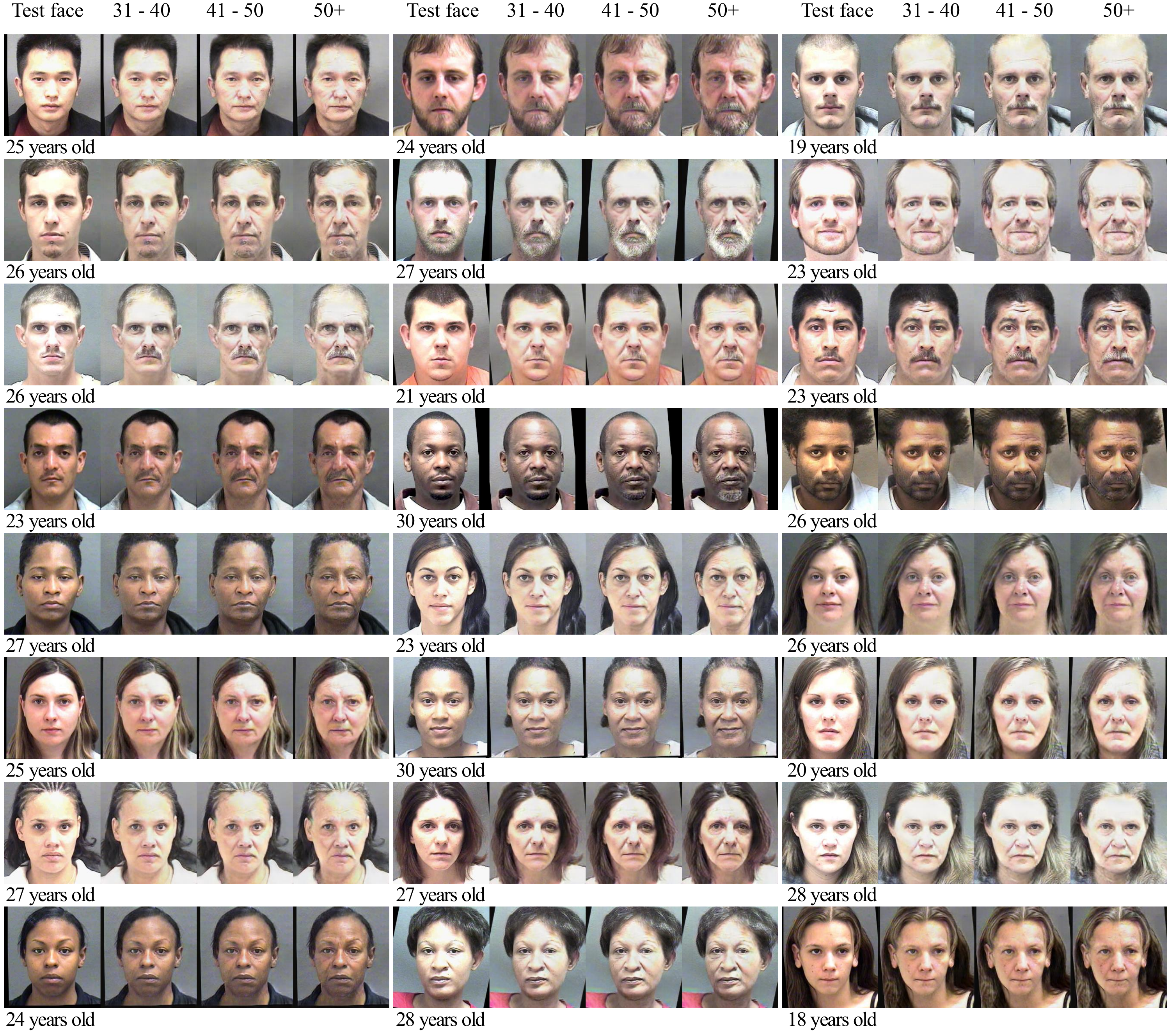}
\vspace{-0.3cm}
\caption{Aging effects obtained on the MORPH databases for 24 different subjects.} 
\vspace{-0.2cm}
\end{figure*}

\begin{figure*}[t]
\centering 
\includegraphics[width=1\textwidth]{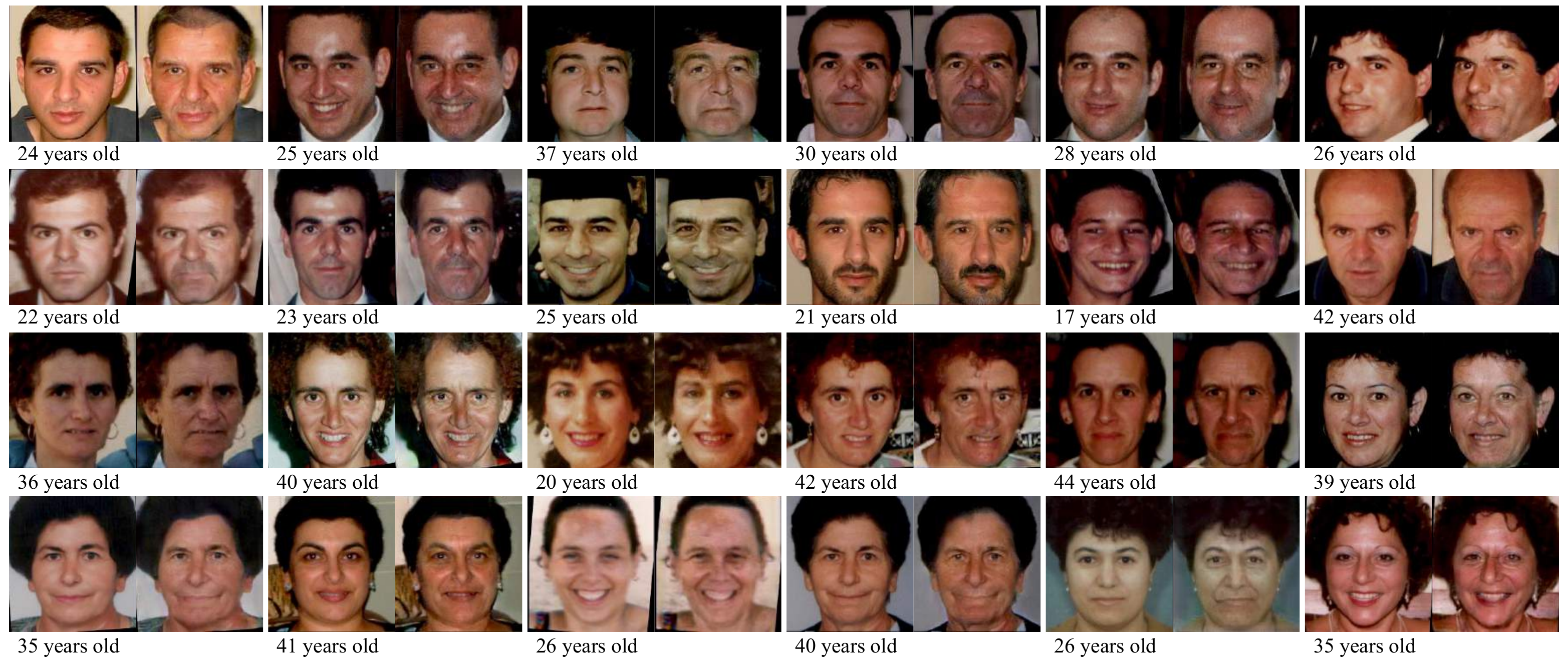}
\vspace{-0.4cm}
\caption{Cross-dataset aging results achieved on the FG-NET dataset for 24 different faces. The first image in each panel is the input face image and the second is the corresponding aging result in the 50+ age cluster.}
\end{figure*}

\begin{figure*}[t]
\centering 
\includegraphics[width=1\textwidth]{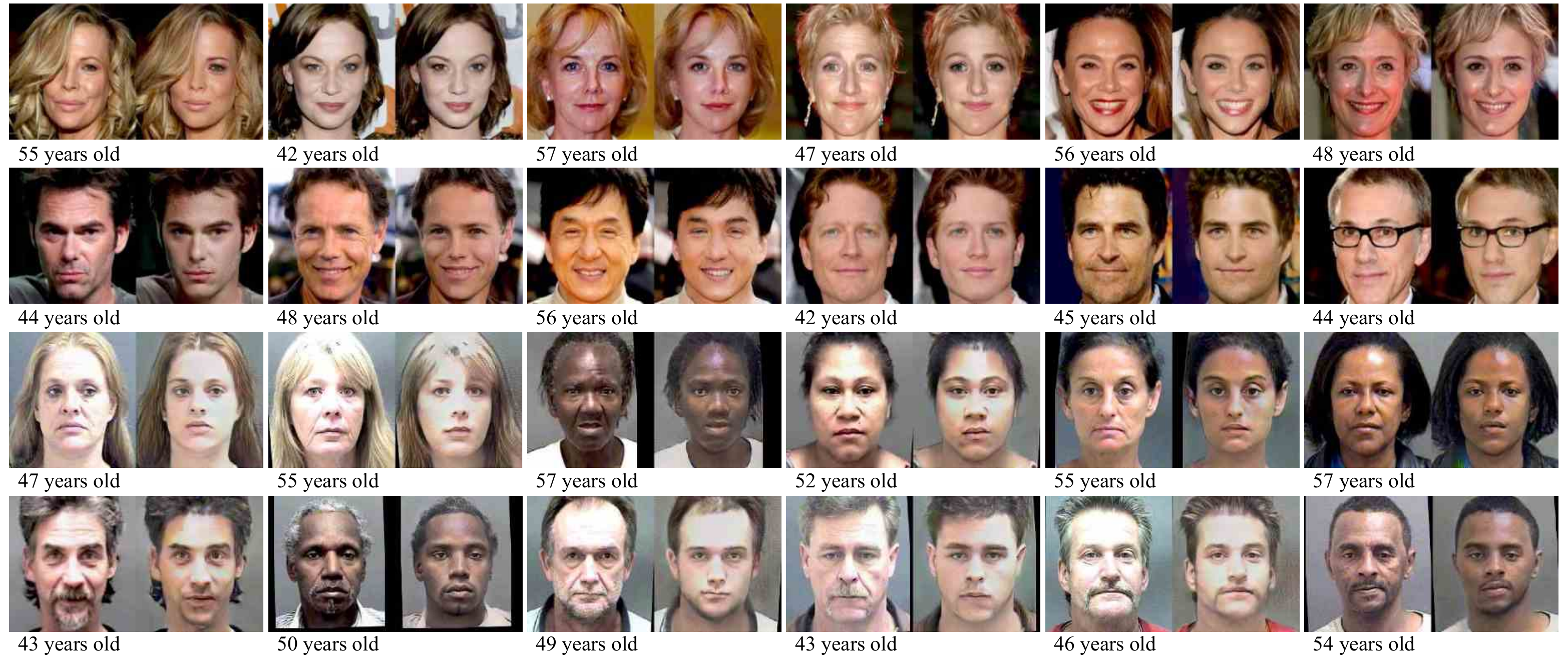}
\vspace{-0.3cm}
\caption{Rejuvenating results achieved on CACD (the top two rows) and MORPH (the bottom two rows) for 24 different subjects. The first image in each panel is the original face image and the second is the corresponding rejuvenating result.}
\end{figure*}

\begin{figure*}[t]
\centering 
\includegraphics[width=1\textwidth]{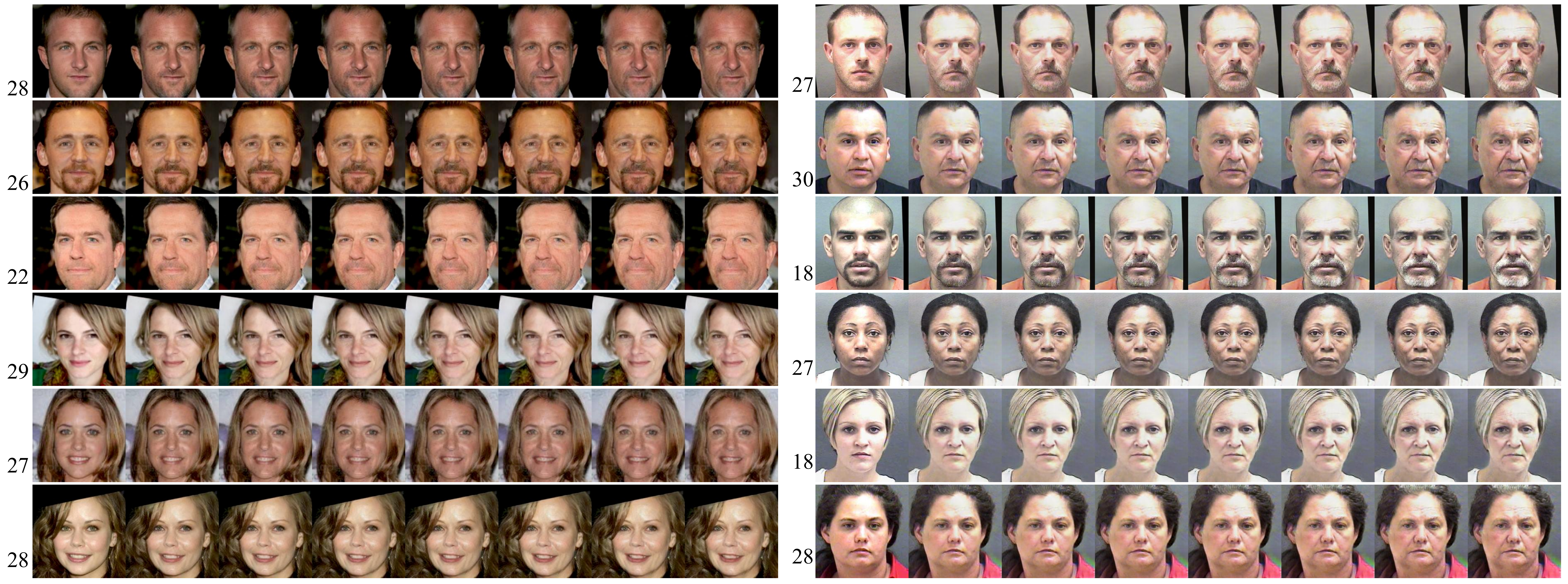}
\caption{Examples of continuous face aging sequences. In each row the leftmost image is the input; and the second, the fifth, and the rightmost images are the results conditioned on the discrete age labels presented to the network during training, while others are interpolated results.}
\vspace{-0.25cm}
\end{figure*}

\begin{figure}[h]
\centering
\includegraphics[width=3.4in]{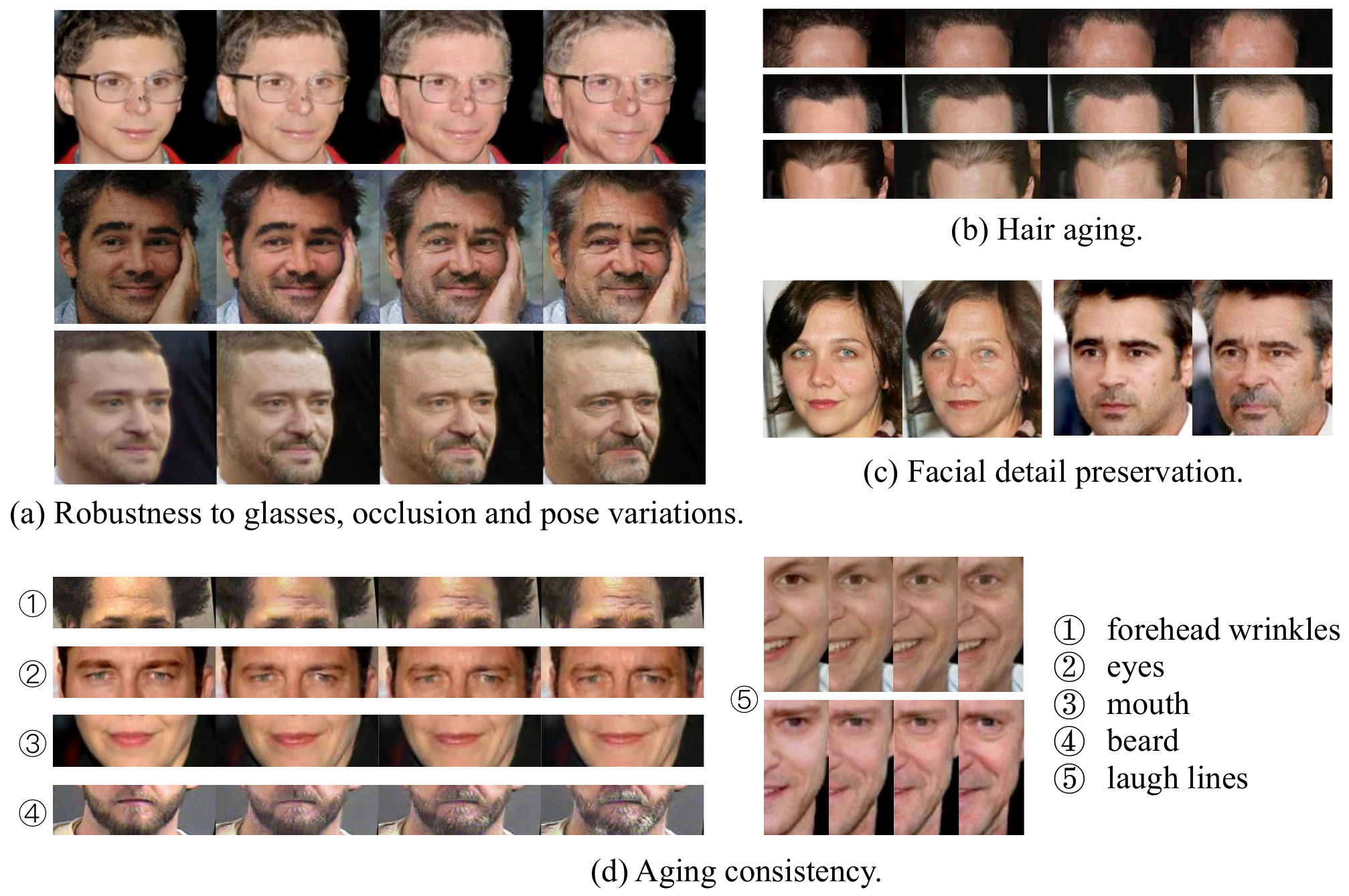}
\caption{Illustration of visual fidelity (zoom in for a better view).}
\end{figure}

\subsection{Results}
\subsubsection{\textbf{Experiment I: Aging Effect Simulation}}

\textbf{Experiment I-A: Discrete Age Progression.}
Five-fold cross validation is conducted to simulate aged faces. On CACD, each fold contains 400 individuals with nearly 10,079,  ~8,635,  ~7,964, and ~6,011 face images from the four age clusters of [14-30], [31-40], [41-50], and [51-60] years, respectively; while on MORPH, each fold consists of around 2,586 subjects with 4,467, ~3,030, ~2,205, and 639 faces from the four age groups. For each run, four folds are utilized for training, and the remainder for evaluation. Examples of age progression results achieved on the two databases are depicted in Figs. 7 and 8. Additionally, cross-dataset validation is conducted on the faces older than 14 years old on FG-NET, with CACD as the training set, and the simulation results are shown in Fig. 9. As we can see, although the examples cover a wide range of population in terms of race, gender, pose, makeup and expression, visually plausible and convincing aging effects are achieved.

Apart from face age progression, the proposed method can be applied for age regression as well. All the test faces in this experiment come from the people older than 30 years old, and they are transformed to the age bracket of below 30 years old. Under such settings, only one discriminator is exploited at the training stage. Example rejuvenating visualizations are shown in Fig. 10. As expected, this operation tightens the face skin, and the hair becomes thick and luxuriant.

\textbf{Experiment I-B: Continuous Age Progression.}
Recall that the proposed aging method can not only generate faces at specific age range presented to the network during training, but also fill up the intermediate transitional states, producing very smooth aging sequences. It indicates that the generator does not simply remember the aging templates, but internally learns meaningful face representation in a latent space, thus able to understand the connections between age clusters.

To be specific, the well-trained aging models in Experiment I-A are directly used for testing in this experiment. The conditional age is first denoted as a tensor of size $224 \times 224 \times k$, and the i$\textit{th}$ channel is set to ones while the others are zeros when the target is the i$\textit{th}$ age cluster. To bridge the gap between discrete age clusters and obtain continuous aged renderings, the item values in the tensor gradually fall and rise within the interval [0,1] at the testing phase, ensuring that the conditional age label smoothly shifts from an existing one to another.

Some representative examples of continuous aging sequences are shown in Figure 11. The images in each panel are sorted by the increasing conditional age. The leftmost image is the input; and the second, the fifth, and the rightmost images are the results conditioned on the existent age labels, while others are interpolated results. As shown in the figure, all the generated images are natural and with high quality, clearly highlighting the ability of knowledge transfer. The aging changes between neighboring images are inconspicuous, while they are convincing throughout the complete aging sequence. Even if only a specific number of discriminators are used and the given age distributions are independently presented to the network during training, the method still successfully and flexibly steers age transformation to any arbitrary age label, making it more useful in the real world.

\subsubsection{\textbf{Experiment II: Aging Model Evaluation}}
We acknowledge that face age progression is supposed to aesthetically predict the future appearance of the individual, beyond aging accuracy and identity preservation, therefore in this experiment a more comprehensive evaluation of the age progression results is provided with both the visual and quantitative analysis.

\textbf{Experiment II-A: Visual Fidelity.}
Fig. 12 (a) displays example face images with glasses, occlusions, and pose variations. The age-progressed faces are still photorealistic and true to the original inputs; whereas the previous prototyping based methods \cite{wang2006age}\cite{Suo:Compositional} are inherently inadequate for such circumstances, and the parametric aging models \cite{Suo:Concatenational}\cite{shu2015personalized} may lead to ghosting artifacts. In Fig. 12 (b), some examples of hair aging are demonstrated. As far as we know, almost all aging approaches proposed in the literature \cite{Kemelmacher:aging}\cite{shu2015personalized}\cite{Yang:faceAging}\cite{Wang:Recurrent}\cite{zhang2017age}\cite{liu2017aging} focus on cropped faces without considering hair aging, mainly because hair is not as structured as the face area. Further, hair is diverse in texture, shape, and color, thus difficult to model. Nevertheless, the proposed method takes the whole face as input, and the hair grows wispy and thin in aging simulation. Fig. 12 (c) confirms the capability of preserving the necessary facial details during aging, \textit{e.g.}, the skin pigmentation, and Fig. 12 (d) shows the smoothness and consistency of the aging changes, where the lips become thinner, the under-eye bags become more obvious, and wrinkles are deeper.

\textbf{Experiment II-B:  Aging Accuracy.} 
Along with face aging, the estimated age is supposed to increase. Correspondingly, objective age estimation is conducted to measure the aging accuracy. We apply the online face analysis tool of Face++\footnote{All the evaluation results from Face++ were obtained on Nov. 2018. The system  is updated at irregular intervals.} \cite{facepp} to every synthesized face on CACD and MORPH. Excluding those undetected, the age-progressed faces of $50,148$ test samples in the CACD dataset are investigated (average of 10,030 test faces in each run under 5-fold cross validation). Table 5 shows the results, where the mean values are 40.52, 48.03, and 54.05 years old for the 3 designated age clusters, respectively. Ideally, they would be observed in the age range of [31-40], [41-50], and [51-60]. Admittedly, the lifestyle factors may accelerate or slow down the aging rates for the individuals and makeups would also influence the appearances, leading to deviations of the estimated age from the actual age, but the overall trend is relatively robust. Due to such intrinsic ambiguities, objective age estimation is further conducted on all the true face images in the dataset as benchmark. In Table 5 and Figs. 13 (a) and 13 (c), it can be seen that the estimated ages of the synthesized faces are well matched with those of the real images, and the estimated ages increase steadily with the elapsed time, clearly validating our method. 

On MORPH, the aging synthesis results of $22,319$ faces below 30 years old are used in this evaluation (average of 4,464 test faces in each run), and it shows similar results, which confirms that the proposed method has indeed captured the data density of the given subset of faces in terms of age. See Table 5 and Figs. 13 (b) and 13 (d) for detailed results.

\textbf{Experiment II-C: Identity Preservation.}
Objective face verification with Face++ is carried out to quantitatively determine if the original identity property is well preserved during age progression. For each test face, we perform comparisons between the input image and the corresponding aging simulation results: [test face, aged face 1], [test face, aged face 2], and [test face, aged face 3]; and statistical analyses among the synthesized faces are conducted, {\em i.e.} [aged face 1, aged face 2], [aged face 1, aged face 3], and [aged face 2, aged face 3]. Similar to experiment II-B, 50,148 young faces in CACD and their age-progressed renderings are used in this evaluation, leading to a total of $50,148 \times 6$ verifications. As shown in Table 6, the obtained verification rates for the three age-progressed clusters are 100 $\pm$ 0 \%, 100 $\pm$ 0 \%, and 99.98 $\pm$ 0.02 \%, respectively. For MORPH, there are $22,319\times 6$ verifications in total, and the verification rates under the five-fold cross validation scheme are 100 $\pm$ 0 \%, 100 $\pm$ 0 \%, and 99.88 $\pm$ 0.07 \%, respectively. Compared to our previous work on face age progression where the aging models are independently trained for each target age cluster\footnote{We re-evaluate the aging results with the latest version of Face++. On CACD, the mean verification rates for the three age clusters are 99.99\%, 99.95\%, and 99.11\%, respectively; and for MORPH, they are 100\%, 99.44\%, and 95.94\%, respectively.}\cite{yang2018AgeProgression}, our work makes remarkable progress for preserving the identity information. It highlights the reliability of the proposed method and validates the necessity of smoothing the transitional states. Additionally, in Table 6 and Fig. 14, face verification confidence decreases as the time elapsed between two images increases, which conforms to the physical effect of face aging \cite{best2018longitudinal}\cite{Deb2017Face}. It may also explain the better performance achieved on CACD in this evaluation, where the maximum mean age gap between the input and synthesized age cluster is 23.09 years, far less than that of 28.61 years achieved on MORPH.

\textbf{Experiment II-D: Contribution of Pyramid Architecture.}
One model assumption is that the pyramid structure of the discriminator $D$ advances the generation of the aging effects, making the age-progressed faces more natural. Accordingly, we conduct ablation study and carry out comparison to the one-pathway discriminator, under which the generated faces are directly fed into the estimator rather than represented as feature pyramid first. The discriminator architecture in the contrast experiment is equivalent to a chaining of the network $\phi_{age}$ and the first pathway in the proposed pyramid $D$.

Fig. 15 provides a demonstration. Visually, the synthesized aging details of the counterpart are not so evident, and the proposed method behaves better in the relatively complex situation, \textit{e.g.} rejuvenating the white beard. To make the comparison more specific and reliable, quantitative evaluations are further conducted with the same settings as in experiments II-B and II-C, and the statistical results are shown in Table 7. In the table, the estimated ages achieved on CACD and MORPH are generally higher than the benchmark (except for the 1st age cluster for CACD), and the mean absolute errors over the three age clusters are 2.48 and 2.67 years for the two databases, respectively, exhibiting larger deviations than 0.99 and 0.85 years obtained by using the pyramid architecture. The reason lies in that the synthesized wrinkles in this contrast experiment are not so clear and the faces look relatively messy. It also explains the decreased face verification confidence observed in Table 7 in the identity preservation evaluation. Based on both the visual and quantitative analysis, we can draw an inference that compared with the pyramid architecture, the one-pathway discriminator, as widely utilized in previous GAN-based frameworks, is lagging behind in regard to modeling the sophisticated aging changes.

\begin{table*}[t] \footnotesize
\centering
\caption{Objective age estimation results (in years) on CACD and MORPH}
\label{faceverification}
\begin{threeparttable}
\begin{tabular}{ccccccccc}
 \multicolumn{4}{c}{  CACD } &  &\multicolumn{4}{c}{ MORPH}  \\
  \textbf{Age Cluster 0}  &\textbf{Age Cluster 1}  & \textbf{Age Cluster  2} &  \textbf{Age Cluster  3} &  &\textbf{Age Cluster 0}  & \textbf{Age Cluster  1}  & \textbf{Age Cluster  2} &  \textbf{Age Cluster  3}  \\  
\midrule
 \multicolumn{4}{c}{ Synthesized faces\tnote{*}} &  &\multicolumn{4}{c}{Synthesized faces\tnote{*}}  \\
\cline{2-4}
\cline{7-9}
\specialrule{0em}{1pt}{1pt}
 \text{--} & 40.52 $\pm$ 9.08 & 48.03 $\pm$ 9.32 & 54.05 $\pm$ 9.94   & & \text{--} &  39.62 $\pm$ 7.29 & 48.09 $\pm$ 7.01  & 56.54 $\pm$ 6.74 \\
\text{--} &  40.52 $\pm$ 0.16 & 48.03 $\pm$ 0.32 & 54.05 $\pm$ 0.23 & & \text{--} &  39.62 $\pm$ 0.84 & 48.09 $\pm$ 1.07  & 56.54 $\pm$ 1.19\\
 \multicolumn{4}{c}{Natural faces}  &  &\multicolumn{4}{c}{Natural faces}  \\
\cline{1-4}
\cline{6-9}
\specialrule{0em}{1pt}{1pt}
 30.96 $\pm$ 8.50 & 38.92 $\pm$ 9.73 &46.95 $\pm$ 10.70 & 53.75 $\pm$ 12.45 & &   27.93 $\pm$ 6.16 & 38.87 $\pm$ 7.52 &  48.03 $\pm$ 8.32 & 58.29 $\pm$ 8.76 \\
\bottomrule
\end{tabular}
\begin{tablenotes}
\footnotesize
\item[*] The standard deviation in the first row is calculated on all the synthesized faces; the standard deviation in the second row is calculated on the mean values of the 5 folds.
\end{tablenotes}     
\end{threeparttable}
\end{table*}

\begin{table*}[t] \footnotesize
\centering\caption{Objective face verification results on (a)~CACD and (b)~MORPH}
\label{faceverification}
\begin{threeparttable}
\begin{tabular}{cccccccccc}
& & \textbf{Aged face 1}  & \textbf{Aged face 2} &  \textbf{Aged face 3} & ~& & \textbf{Aged face 1}  & \textbf{Aged face 2} &  \textbf{Aged face 3}  \\  
\midrule
~& &\multicolumn{3}{c}{verification confidence\tnote{a}  } & ~& &\multicolumn{3}{c}{verification confidence\tnote{a} }  \\
\cline{3-5}
\cline{8-10} 
\specialrule{0em}{1pt}{1pt}
\textbf{Test face} &  \multirow{10}{*}{(a)}  & 94.61 $\pm$ 0.07 & 93.13 $\pm$ 0.24 & 91.22 $\pm$ 0.25   & &\multirow{10}{*}{(b)}  &     94.65 $\pm$ 0.11 & 92.46 $\pm$ 0.23 & 88.12 $\pm$ 0.46 \\
\textbf{Aged face 1} & & \text{--}  &    96.71 $\pm$ 0.02 & 95.45 $\pm$ 0.06   & & & \text{--}   & 96.59 $\pm$ 0.04 & 94.18 $\pm$ 0.11  \\
\textbf{Aged face 2} & & \text{--}  &  \text{--} & 96.71 $\pm$ 0.03   & & &  \text{--}   &  \text{--}   & 96.20 $\pm$ 0.06\\
~ & & \multicolumn{3}{c}{verification confidence \tnote{b} }  & ~& &\multicolumn{3}{c}{verification confidence\tnote{b} }  \\
\cline{3-5}
\cline{8-10} 
\specialrule{0em}{1pt}{1pt}
\textbf{Test face} & &  94.61 $\pm$ 1.00 & 93.13 $\pm$ 1.68 & 91.22 $\pm$ 2.55 & & &    94.65 $\pm$ 0.95 & 92.46 $\pm$ 1.87 & 88.12 $\pm$ 3.30\\
\textbf{Aged face 1} & & \text{--}  & 96.71 $\pm$ 0.29 & 95.45 $\pm$ 0.79& &  &\text{--}   &96.59 $\pm$ 0.27 & 94.18 $\pm$ 1.14 \\
\textbf{Aged face 2}  & & \text{--}  &  \text{--} & 96.71 $\pm$ 0.26 & &  &\text{--}   &  \text{--}   & 96.20 $\pm$ 0.40 \\
~ & & \multicolumn{3}{c}{verification rate (threshold = 73.98, FAR = 1e - 5)}  & ~ & &\multicolumn{3}{c}{verification rate (threshold = 73.98, FAR = 1e - 5)}  \\
\cline{3-5}
\cline{8-10} 
\specialrule{0em}{1pt}{1pt}
\textbf{Test face} & & 100 $\pm$ 0 \% & 100 $\pm$ 0 \% &  99.98 $\pm$ 0.02 \%  & & & 100 $\pm$ 0 \% & 100 $\pm$ 0 \% & 99.88 $\pm$ 0.07 \%\\
\bottomrule
\end{tabular}
\begin{tablenotes}
\footnotesize
\item[a] The standard deviation is calculated on the mean values of the 5 folds.
\item[b] The standard deviation is calculated on all the synthesized faces. 
\end{tablenotes}     
\end{threeparttable}
\end{table*}

\begin{figure}[t] 
\centering
\includegraphics[width=3.5in]{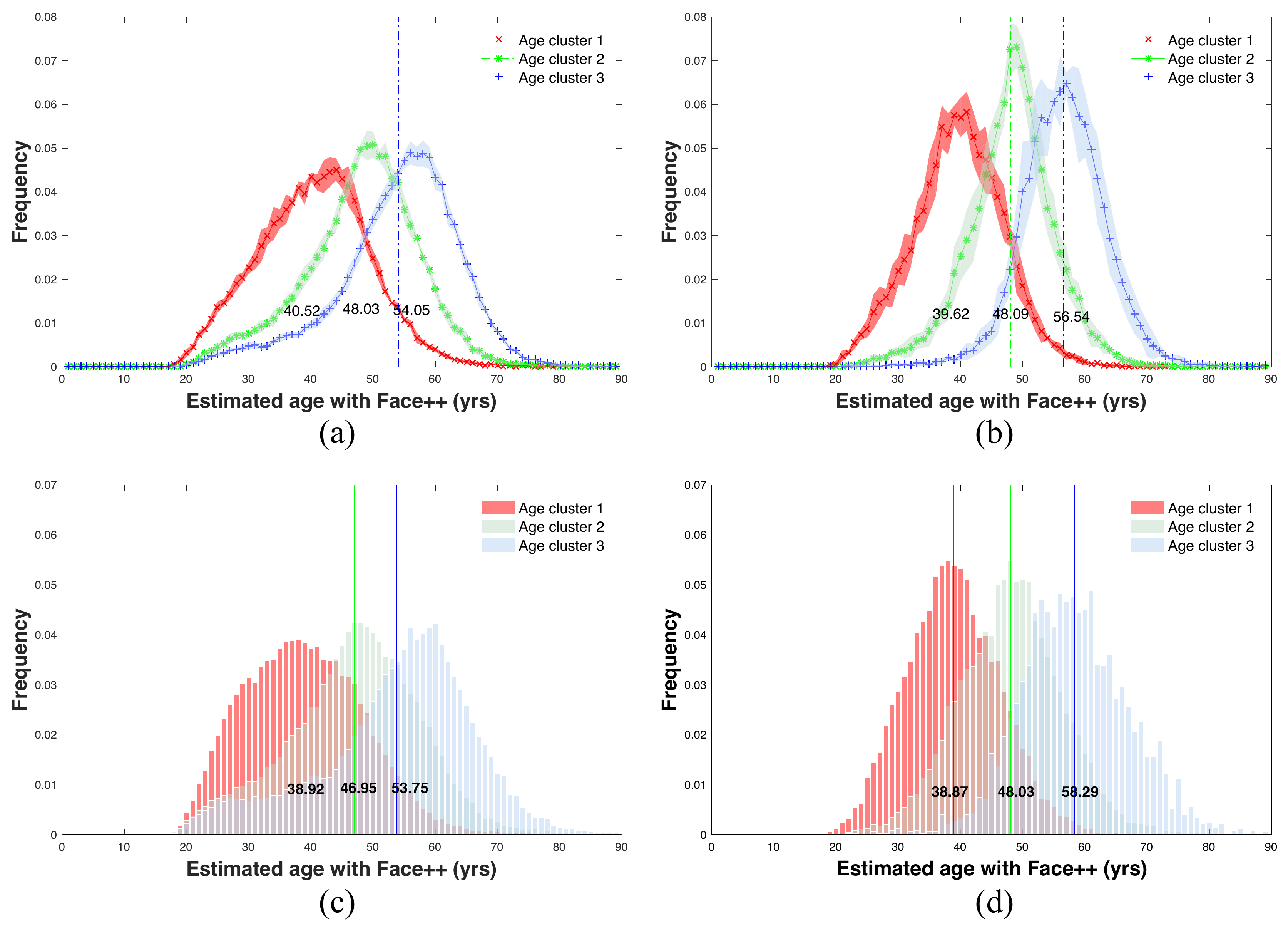}
\caption{Distributions of the estimated ages obtained by Face++. (a) CACD, synthesized faces; (b) MORPH, synthesized faces; (c) CACD, actual faces; and (d) MORPH, actual faces.} 
\vspace{-0.3cm}
\end{figure}

\begin{figure}[t] 
\centering
\includegraphics[width=3.5in]{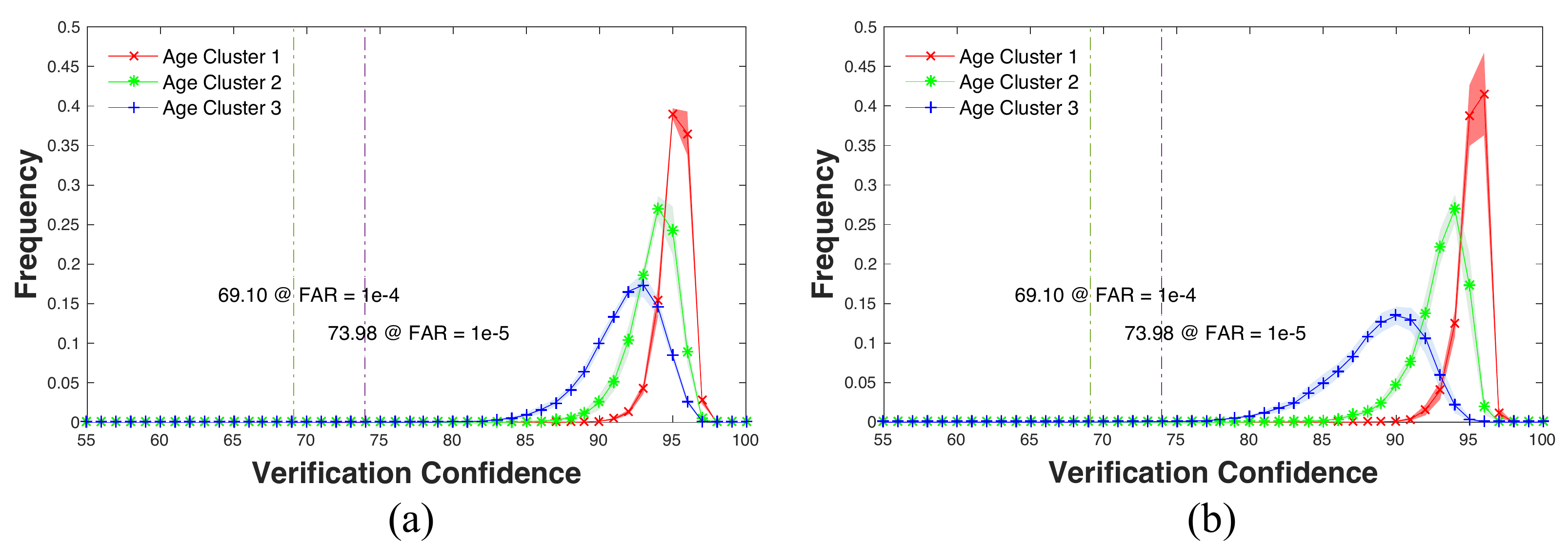}
\caption{Distributions of the face verification confidence on (a) CACD and (b) MORPH.} 
\end{figure}

\textbf{Experiment II-E: Comparison to Prior Work.}
To compare with prior work, we conduct testing on the FG-NET and MORPH databases. These prior studies are \cite{park2010age}\cite{Suo:Concatenational}\cite{Yang:faceAging}\cite{shu2015personalized}\cite{nhan2016longitudinal}\cite{Wang:Recurrent}\cite{zhang2017age}\cite{nhan2017temporal}\cite{liu2017aging}, which signify the state-of-the-art. In addition, one of the most popular mobile aging applications, {\em i.e.} \textit{Agingbooth} \cite{agingbooth}, and the online aging tool \textit{Face of the future} \cite{futureface} are also compared. Fig. 16 displays some example faces. As it can be seen, \textit{Face of the future} and \textit{Agingbooth} follow the prototyping-based method, where the identical aging mask is indiscriminately applied to all the given faces as most of the aging Apps do. While the concept of such methods is straightforward, the age-progressed faces are not photorealistic. Regarding the published works in the literature, ghosting artifacts are unavoidable for the parametric method \cite{Suo:Concatenational} and the dictionary reconstruction based solutions \cite{Yang:faceAging}\cite{shu2015personalized}. Technological advancements can be observed in the deep generative models \cite{Wang:Recurrent}\cite{zhang2017age}\cite{liu2017aging}, whereas they only focus on the cropped facial area, and the age-progressed faces lack necessary aging details. In a further experiment, we collect 138 paired images of 54 individuals from the published papers, and invite 10 human observers to evaluate which age-progressed face is better in the pairwise comparison. Among the 1,380 votes, 71.74\% prefer the proposed method, 19.28\% favor the prior work, and 8.98\% indicate that they are about the same. Besides, the proposed method does not require burdensome preprocessing as previous works do, and it only needs 2 landmarks for pupil alignment. To sum up, we can say that the proposed method outperforms the counterparts.

\begin{table*}[t] \footnotesize
\centering
\caption{Quantitative evaluation results using one-pathway discriminator on (a) CACD and (b) MORPH.}
\label{FaceVerification}
\begin{tabular}{cccccccccc}
\toprule
& & \textbf{Aged face 1}  & \textbf{Aged face 2} &  \textbf{Aged face 3} & ~& & \textbf{Aged face 1}  & \textbf{Aged face 2} &  \textbf{Aged face 3}  \\  
\cline{3-5}
\cline{8-10}
\specialrule{0em}{1pt}{1pt}
 \textbf{Estimated age (yrs old)} & \multirow{2}{*}{(a)}  &  36.74 $\pm$ 9.49 & 48.95 $\pm$ 8.71 & 57.01 $\pm$ 10.11 & &\multirow{2}{*}{(b)}  &  42.32 $\pm$ 8.21 & 51.45 $\pm$ 8.02 & 59.43 $\pm$ 7.75 \\
\specialrule{0em}{1pt}{1pt}
\textbf{Verification confidence} & & 95.51 $\pm$ 0.70   & 91.52 $\pm$ 2.47 & 87.43 $\pm$ 4.27 & &  & 94.00 $\pm$ 1.10  & 91.13 $\pm$ 2.05 & 86.94 $\pm$  3.43 \\
\bottomrule
\end{tabular}
\end{table*}

\begin{figure*}[t] 
\centering
\includegraphics[width=0.95\textwidth]{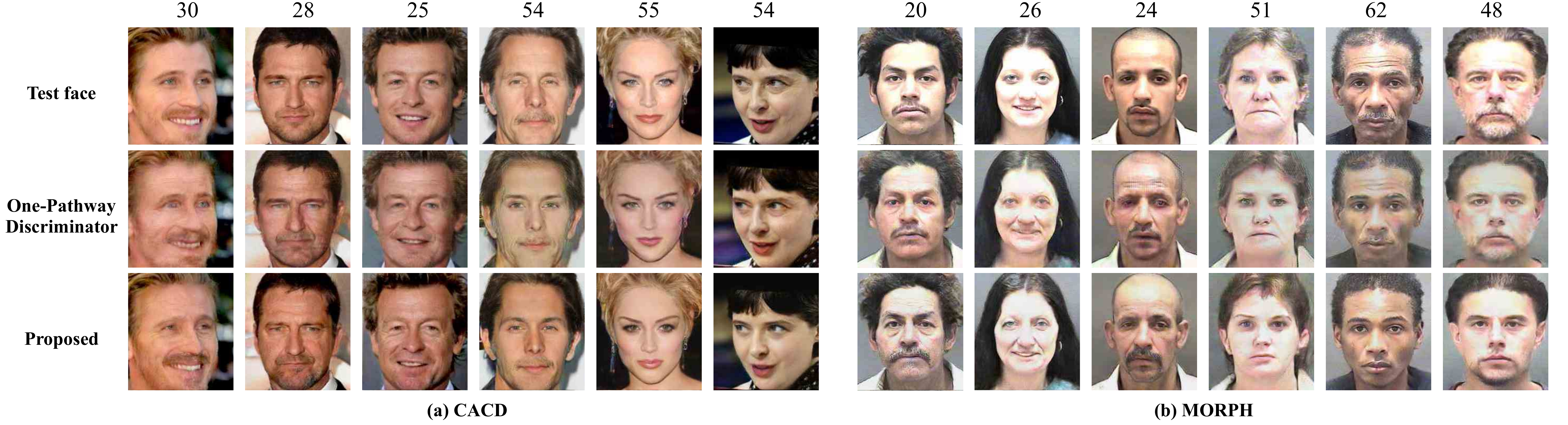}
\caption{Visual comparison to the one-pathway discriminator on (a) CACD and (b) MORPH. For both datasets, the left three columns show the aging results, while the right three show the rejuvenating results.}
\end{figure*}

\begin{figure*}[t] 
\centering
\includegraphics[width=6.47in]{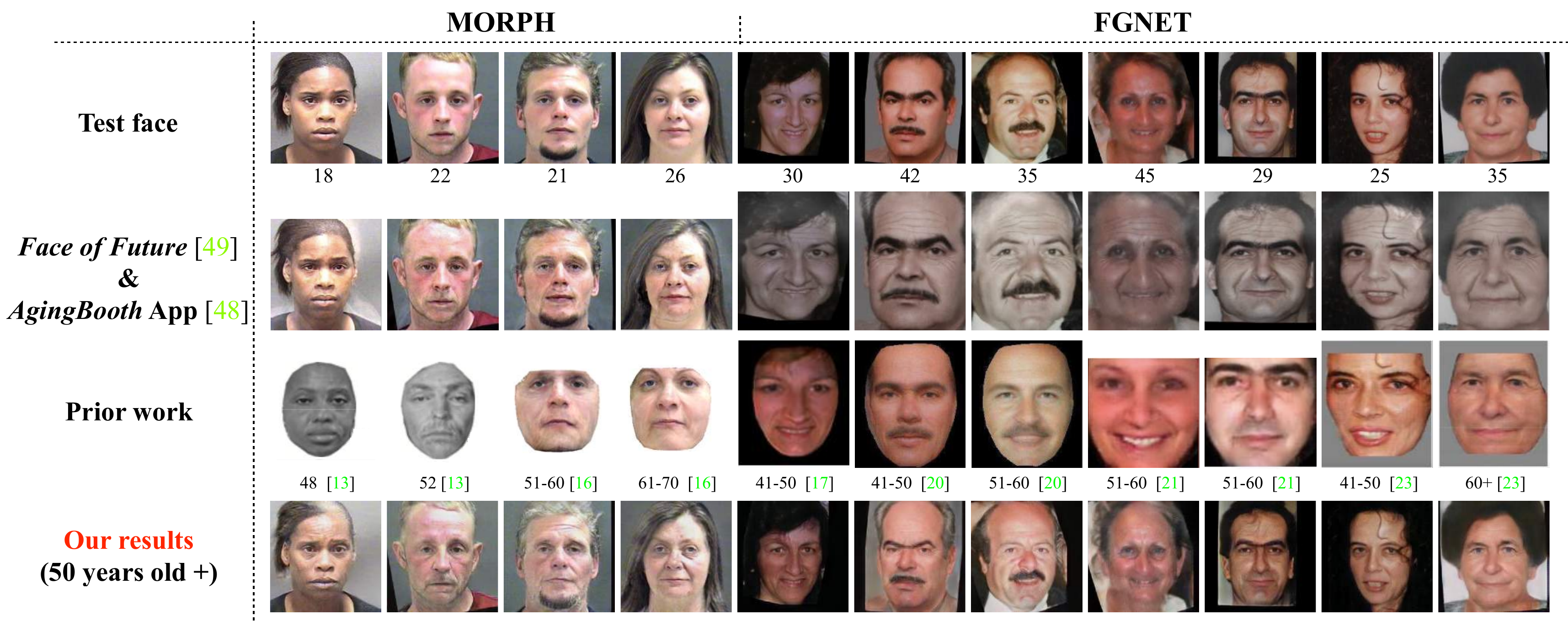}
\caption{Performance comparison with prior work (zoom in for a better view of the aging details).}
\vspace{-0.2cm}
\end{figure*}

\section{Conclusion}
This study presents an effective solution to aging accuracy and identity preservation, and proposes a novel GAN based method. It exploits a compound training critic that integrates the simple pixel-level penalty, the age-related GAN loss achieving age transformation, and the individual-dependent critic keeping the identity information stable. For generating detailed signs of aging, a pyramidal architecture of discriminator is designed to estimate high-level face representations in a finer way. An adversarial learning scheme is further presented, to simultaneously train a single generator and multiple parallel discriminators, enabling the model to generate smooth continuous face aging sequences. Extensive experiments are conducted on three datasets, and the proposed method is shown to be effective in generating diverse face samples. Quantitative evaluations from a COTS face recognition system show that the target age distributions are accurately recovered; and 99.88\% and 99.98\% age progressed faces can be correctly verified at $0.001\%$ FAR after age transformations of approximately 28.61 years elapsed time on MORPH and 23.09 years on CACD.

The proposed approach achieves more accurate, more reliable, and more photorealistic aging effects than the state of the art. But, it indeed has some limitations. On the one hand, we primarily consider the general aging procedure and the facial properties that are inextricably bound to identity. There are actually additional covariates of interest that could largely influence face aging while cannot be taken into account, \textit{e.g.} health condition, living style, and working environment, mainly due to the inaccessibility of such information. On the other hand, a solution to adult aging is provided, and child growth is given less attention. A shortage of publicly available longitudinal face dataset of children \cite{Best-Rowden2016} is partly responsible for this, while another reason lies in that the identity feature of the younger individuals are less stable. As remarked by the recent findings on human face recognition conducted by NIST\cite{nist2014}, children are not easy to recognize, which might make aged renderings questionable. Both the above-mentioned unsolved issues could be the major directions in the future work.


%





%

\begin{IEEEbiography}[{\includegraphics[width=1in,height=1.25in,clip,keepaspectratio]{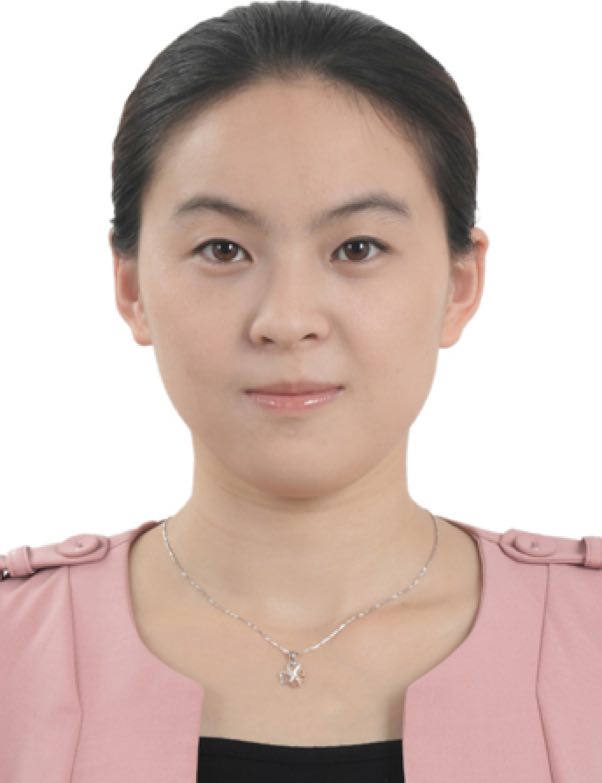}}]{Hongyu Yang}
received the B.E. degree in computer science and technology from Beihang University, Beijing, China, in 2013, where she is currently working toward the Ph.D. degree with the Laboratory of Intelligent Recognition and Image Processing. From 2016 to 2017, she was a visiting scholar at Michigan State University, USA. Her research interests include age invariant face recognition, face aging synthesis, and age estimation.
\end{IEEEbiography}

\begin{IEEEbiography}[{\includegraphics[width=1in,height=1.25in,clip,keepaspectratio]{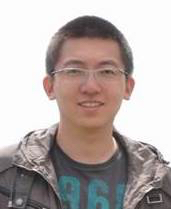}}]{Di Huang}
received the B.S. and M.S. degrees in computer science from Beihang University, Beijing, China, and the Ph.D. degree in computer science from the \'Ecole centrale de Lyon, Lyon, France, in 2005, 2008, and 2011, respectively. He joined the Laboratory of Intelligent Recognition and Image Processing, School of Computer Science and Engineering, Beihang University, as a Faculty Member. His current research interests include biometrics, in particular, on 2D/3D face analysis, image/video processing, and pattern recognition.
\end{IEEEbiography}

\begin{IEEEbiography}[{\includegraphics[width=1in,height=1.25in,clip,keepaspectratio]{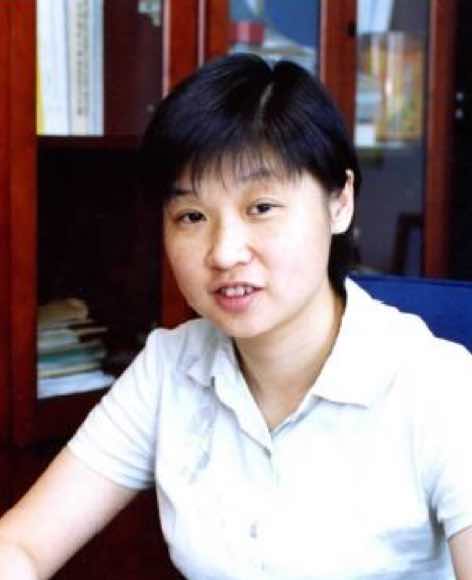}}]{Yunhong Wang}
received the B.S. degree in electronic engineering from Northwestern Polytechnical University, Xi'an, China, in 1989, and the M.S. and Ph.D. degrees in electronic engineering from Nanjing University of Science and Technology, Nanjing, China, in 1995 and 1998, respectively. She was with the National Laboratory of Pattern Recognition, Institute of Automation, Chinese Academy of Sciences, Beijing, China, from 1998 to 2004. Since 2004, she has been a Professor with the School of Computer Science and Engineering, Beihang University, Beijing, where she is also the Director of Laboratory of Intelligent Recognition and Image Processing, Beijing Key Laboratory of Digital Media. Her current research interests include biometrics, pattern recognition, computer vision, data fusion, and image processing.
\end{IEEEbiography}

\begin{IEEEbiography}[{\includegraphics[width=1in,height=1.25in,clip,keepaspectratio]{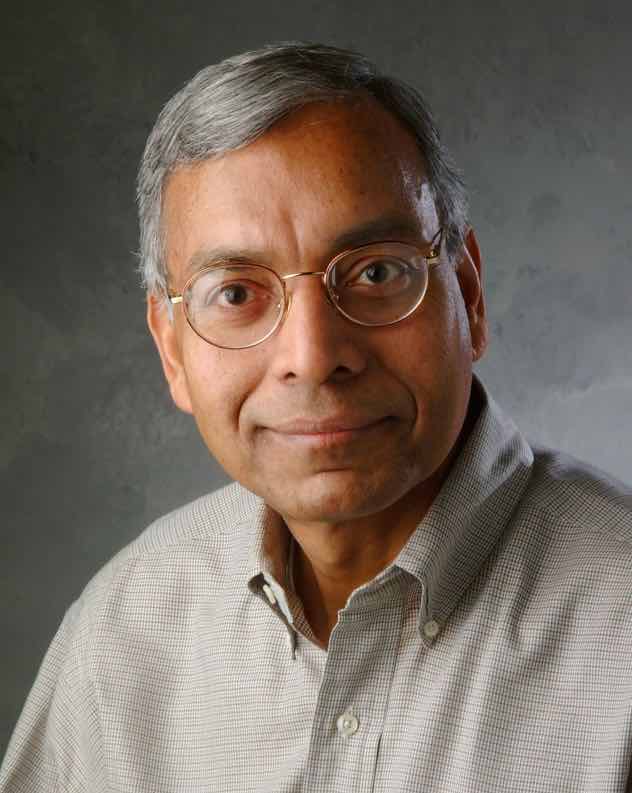}}]{Anil K. Jain}
is a University distinguished professor in the Department of Computer Science and Engineering at Michigan State University. His research interests include pattern recognition and biometric authentication. He served as the editor-in-chief of the IEEE Transactions on Pattern Analysis and Machine Intelligence and was a member of the United States Defense Science Board. He has received Fulbright, Guggenheim, Alexander von Humboldt, and IAPR King Sun
Fu awards. He is a member of the National Academy of Engineering and foreign fellow of the Indian National Academy of Engineering.
\end{IEEEbiography}




\end{document}